\PassOptionsToPackage{numbers,compress,sort}{natbib}
\documentclass[10pt,a4paper,logo]{paper}
\usepackage{natbib}
\usepackage{fontawesome5}
\definecolor{title_blue}{HTML}{204899}
\usepackage{hyperref}       
\usepackage{url}            
\usepackage{booktabs}       
\usepackage{amsfonts}       
\usepackage{nicefrac}       
\usepackage{microtype}      
\usepackage[table]{xcolor}  
\usepackage{graphicx}       
\usepackage{capt-of}        
\usepackage{wrapfig}        

\usepackage{amsmath}
\usepackage{amssymb}
\usepackage{pifont}
\usepackage{mathtools}
\usepackage{amsthm}
\usepackage{bbm}

\usepackage[capitalize,noabbrev,nameinlink]{cleveref}
\crefname{equation}{Eq.}{Eqs.}
\creflabelformat{equation}{#2\textup{#1}#3}
\crefname{section}{Sec.}{Secs.}
\crefname{figure}{Fig.}{Figs.}
\crefname{table}{Table}{Tables}
\crefname{algorithm}{Alg.}{Algs.}

\crefname{thm}{theorem}{theorems}
\Crefname{thm}{Theorem}{Theorems}
\crefname{prop}{proposition}{propositions}
\Crefname{prop}{Proposition}{Propositions}
\crefname{lem}{lemma}{lemmas}
\Crefname{lem}{Lemma}{Lemmas}
\crefname{cor}{corollary}{corollaries}
\Crefname{cor}{Corollary}{Corollaries}
\crefname{rmk}{remark}{remarks}
\Crefname{rmk}{Remark}{Remarks}
\crefname{definition}{definition}{definitions}
\Crefname{definition}{Definition}{Definitions}

\creflabelformat{equation}{#2\textup{#1}#3}

\definecolor{Red}{rgb}{0.768, 0.054, 0.054}
\definecolor{Blue}{rgb}{0.152, 0.294, 0.925}
\definecolor{Green}{rgb}{0.0, 0.55, 0.35}
\definecolor{hotpink}{rgb}{1.0, 0.41, 0.71}
\definecolor{brown}{rgb}{0.59, 0.29, 0.0}
\definecolor{purple}{rgb}{0.59, 0.44, 0.84}
\definecolor{darkpastelgreen}{rgb}{0.01, 0.75, 0.24}
\definecolor{celestialblue}{rgb}{0.29, 0.59, 0.82}
\definecolor{ceruleanblue}{rgb}{0.16, 0.32, 0.75}
\definecolor{goldenrod}{rgb}{0.85, 0.65, 0.13}
\definecolor{navyblue}{rgb}{0.0, 0.0, 0.5}
\definecolor{coolgrey}{rgb}{0.55, 0.57, 0.67}
\definecolor{darkseagreen}{rgb}{0.56, 0.74, 0.56}
\definecolor{darkturquoise}{rgb}{0.0, 0.81, 0.82}
\definecolor{citeblue}{HTML}{1668b0}
\definecolor{linkpink}{HTML}{d61313}
\definecolor{urlpink}{HTML}{b534ad}
\definecolor{myblue}{HTML}{EFF6FF}
\definecolor{mypink}{HTML}{FFF1F5}
\definecolor{capblue}{HTML}{DBEAFE}
\definecolor{cappink}{HTML}{FFE1EA}
\definecolor{avggray}{HTML}{F2F2F2}
\newcommand{\rowblue}{\rowcolor{myblue}}
\newcommand{\rowpink}{\rowcolor{mypink}}
\newcommand{\smallhl}[2]{{\setlength{\fboxsep}{1pt}\colorbox{#1}{#2}}}
\hypersetup{
    colorlinks=true,
    citecolor=ceruleanblue,
    linkcolor=ceruleanblue,
    urlcolor=ceruleanblue,
}

\theoremstyle{plain}

\theoremstyle{definition}

\theoremstyle{remark}

\usepackage{array}
\usepackage{multirow}
\usepackage{tabularx}
\usepackage{float}
\usepackage{algorithm}
\usepackage{algorithmic}

\usepackage[most]{tcolorbox}

\newcommand{\modelname}[1]{\textcolor{blue}{MODEL}}

\usepackage{tcolorbox}
\tcbuselibrary{skins}
\tcbuselibrary{listings}
\tcbuselibrary{breakable}
\newcommand{\promptfont}{\fontsize{6.5}{8}\selectfont\ttfamily}
\newcommand{\promptbefore}{\par\smallskip\Needspace{8\baselineskip}\noindent}
\tcbset{prompt splitting/.style={breakable}}
\newtcblisting{promptbox}[1][]{
    enhanced,
    prompt splitting,
    before={\promptbefore},
    title after break={Continued},
    attach boxed title to top left={xshift=1em, yshift=-\tcboxedtitleheight/2},
    coltitle=white,
    listing only,
    boxed title style={sharp corners, frame hidden},
    fonttitle=\ttfamily\footnotesize\bfseries,
    title={#1},
    listing options={
        language={},
        basicstyle=\promptfont,
        breaklines=true,
        breakautoindent=false,
        breakindent=0pt,
        columns=fullflexible,
        showstringspaces=false,
    },
    boxsep=2pt,
    top=2pt,
    bottom=2pt,
}

\newsavebox{\prompttemplatebox}
\newenvironment{prompttemplate}[2][6.5]{%
    \par\addvspace{6pt}%
    \begin{lrbox}{\prompttemplatebox}%
    \begin{minipage}{\linewidth}%
    \tcbset{prompt splitting/.style={unbreakable}}%
    \renewcommand{\promptfont}{\fontsize{#1}{#2}\selectfont\ttfamily}%
    \renewcommand{\promptbefore}{\par\smallskip\noindent}%
}{%
    \end{minipage}%
    \end{lrbox}%
    \ifdim\dimexpr\ht\prompttemplatebox+\dp\prompttemplatebox\relax>\textheight
        \PackageWarningNoLine{prompttemplate}{Prompt template exceeds the page height}%
    \fi
    \Needspace{\dimexpr\ht\prompttemplatebox+\dp\prompttemplatebox\relax}%
    \noindent\usebox{\prompttemplatebox}\par\addvspace{6pt}%
}

\title{\textsc{PREPING}: Zero-Task Agent Memory Construction \\ via Self-Practice}
\reportnumber{}
\author[1]{Yumin Choi}
\author[1]{Sangwoo Park}
\author[1]{Minki Kang}
\author[1,$\dagger$]{Jinheon Baek}
\author[1,2,$\dagger$]{Sung Ju Hwang}
\affil[1]{KAIST}
\affil[2]{DeepAuto.ai}
\correspondingauthor{}
\renewcommand{\reportcontacts}{%
    \begingroup\normalfont\fontsize{9}{13}\selectfont
    \begin{tabular}{@{}p{1em}@{\hspace{0.35em}}l@{}}
        \faEnvelope[regular] & \href{mailto:yuminchoi@kaist.ac.kr,jinheon.baek@kaist.ac.kr,sungju.hwang@kaist.ac.kr}{\textcolor{black}{\texttt{\{yuminchoi, jinheon.baek, sungju.hwang\}@kaist.ac.kr}}} \\
        \faGlobe & \url{https://dozi01.github.io/preping-project-page/}
    \end{tabular}\par
    \endgroup
}

\begin{abstract}
Agent memory is typically constructed either offline from curated demonstrations or online from post-deployment interactions. However, regardless of how it is built, an agent faces a cold-start gap when first introduced to a new environment without any task-specific experience available. In this paper, we study \textit{pre-task memory construction}: whether an agent can build procedural memory before observing any target-environment tasks, using only self-generated synthetic practice. Yet, synthetic interaction alone is insufficient, as without controlling what to practice and what to store, synthetic tasks become redundant, infeasible, and ultimately uninformative, and memory further degrades quickly due to unfiltered trajectories. To overcome this, we present \textsc{Preping}, a proposer-guided memory construction framework. At its core is \textit{proposer memory}, a structured control state that shapes future practice. A \textit{Proposer} generates synthetic tasks conditioned on this state, a \textit{Solver} executes them, and a \textit{Validator} determines which trajectories are eligible for memory insertion while also providing feedback to guide future proposals. Experiments on AppWorld, BFCL v3, and MCP-Universe show that \textsc{Preping} substantially improves over a no-memory baseline and achieves performance competitive with strong playbook-based methods built from offline or online experience, with deployment cost $2.99\times$ lower on AppWorld and $2.23\times$ lower on BFCL v3 than online memory construction. Further analyses reveal that the main benefit does not come from synthetic volume alone, but from proposer-side control over feasibility, redundancy, and coverage, combined with selective memory updates.
\end{abstract}

\begin{document}
\maketitle
\begin{figure}[H]
    \centering
    \begin{minipage}[t]{0.53\linewidth}
        \vspace{0pt}
        \centering
        \includegraphics[width=\linewidth]{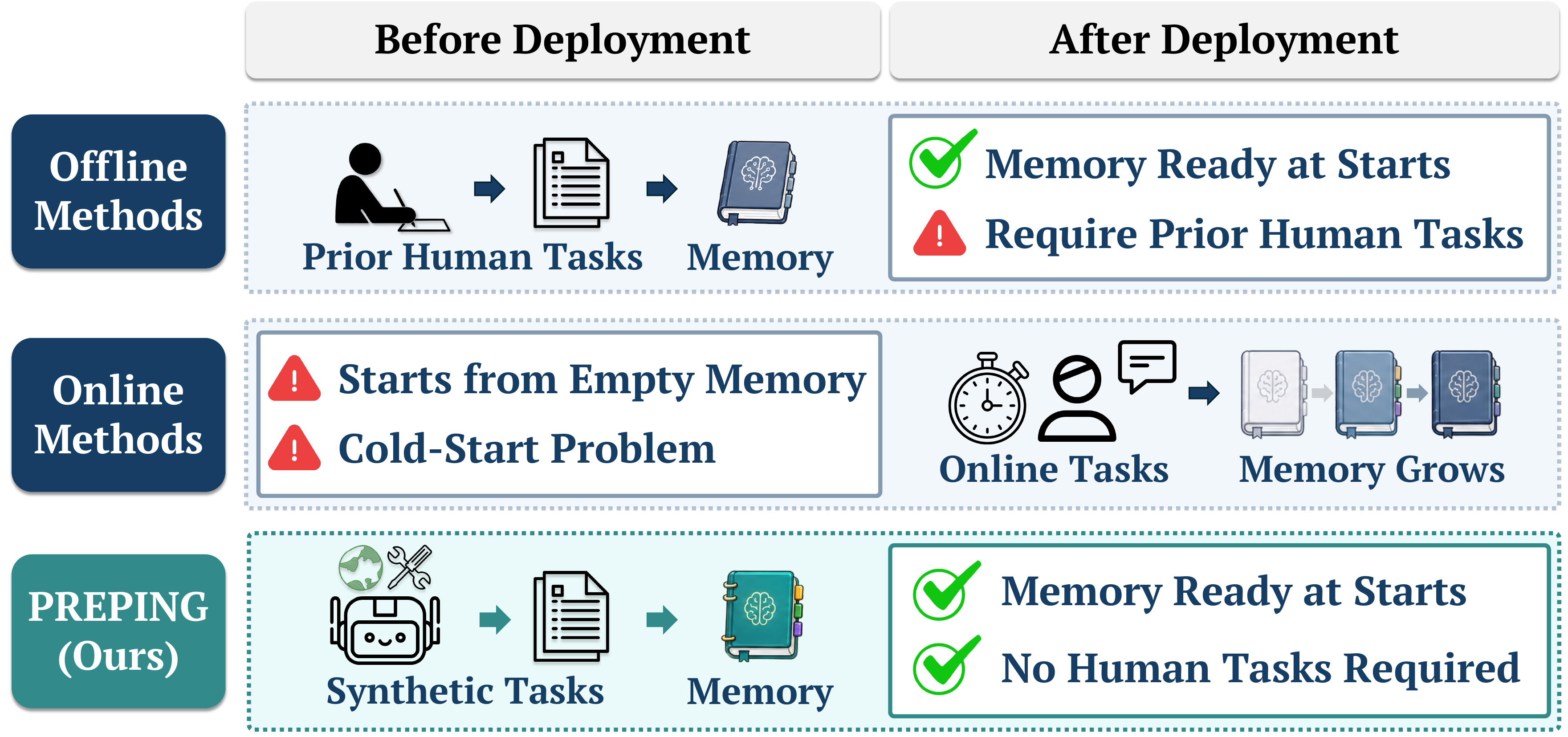}
    \end{minipage}
    \hfill
    \begin{minipage}[t]{0.45\linewidth}
        \vspace{0.05in}
        \centering
        \includegraphics[width=\linewidth]{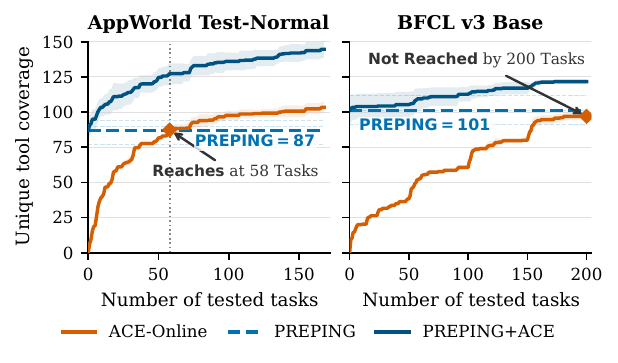}
    \end{minipage}
    \vspace{-0.10in}
    \caption{\textbf{\textsc{Preping} builds memory before the first user task and mitigates online cold-start.}
    Left: Unlike offline methods that require prior human tasks or online methods that start from empty memory, \textsc{Preping} constructs procedural memory through self-generated synthetic practice before deployment.
    Right: \textsc{Preping} establishes broad tool coverage before deployment, whereas ACE-Online must accumulate coverage from user-facing tasks, requiring 58 tasks to match \textsc{Preping} on AppWorld and still falling short after 200 tasks on BFCL v3.}
    \label{fig:concept_fig}
    \label{fig:coverage_cold_start}
\end{figure}

\begingroup
\setlength{\parskip}{4pt plus 1pt minus 0.5pt}
\titlespacing*{\paragraph}{0pt}{9pt plus 1pt minus 1pt}{1em}
\section{Introduction}

LLM agents are increasingly deployed to solve tasks by acting in executable environments, from tool APIs and Model Context Protocol (MCP) servers to command-line interfaces~\citep{react,ace,anthropicmcp,mcpuniverse,terminalbench,officebench}. In these environments, success requires more than knowing which tools are available: agents must learn environment-specific procedures, including how tool calls compose, which preconditions matter, and how to recover from state-dependent failures~\citep{toolformer,gorilla,toolllm}. Reusable memory offers a practical pathway to capture such procedural knowledge as tool-use guidance, workflow rules, and playbook-style instructions. Prior work on agent memory shows that such memory can substantially improve downstream execution performance when useful task experience is available~\citep{dynamiccheatsheet,gmemory,legomem,jefhinter}.

However, existing memory construction methods typically rely on target-environment task experience, either collected before deployment as demonstrations or trajectories, or accumulated after deployment from user interactions~\citep{awm,reasoningbank,memento}. This requirement creates a practical gap for newly connected environments. Offline construction depends on humans to design, collect, or solve tasks per environment, which is costly and rarely available before deployment. Online construction avoids this upfront effort but starts from empty memory: the agent learns only after user-facing tasks arrive, exposing users to early failures, memory-update latency, and additional deployment-time costs. Memory is thus most needed precisely when the experience required to build it has not yet been collected.

Motivated by this, we study \emph{pre-task memory construction}: building reusable procedural memory before any target-environment task data is available. In this setting, the agent may inspect environment documentation, execute tools, and observe their outputs, but it has not seen human-provided tasks, demonstrations, solved trajectories, or deployment-time user interactions from the target environment. This makes the setting distinct from simply performing online memory construction earlier: without task instructions, the agent lacks a direct signal about which user goals will appear, which tools should be composed, or what successful task-level workflows should look like. 

Pre-task memory construction is challenging because access to the environment alone does not reveal reusable task-level procedures. Tool documentation and schemas specify callable interfaces, but often leave environment-specific preconditions, state-dependent constraints, and failure recovery strategies implicit; likewise, free-form exploration can reveal isolated tool-execution examples, but does not reliably produce reusable procedures for accomplishing task-level goals. The agent must therefore create and execute its own task-level objectives through self-generated synthetic practice. However, naive synthetic practice introduces two coupled control problems: tasks can be redundant, infeasible, or poorly grounded in the environment, and storing their trajectories can contaminate memory with misleading guidance. Therefore, pre-task memory construction is not merely a synthetic task generation problem, but a problem of jointly shaping \emph{what to practice} and \emph{what to store}.

To address this, we introduce \textsc{Preping} (\textbf{Pre}-Task \textbf{RE}usable \textbf{P}laybook Mak\textbf{ING}), a framework that couples proposer-guided synthetic practice with validation-gated memory admission. \textsc{Preping} maintains \emph{proposer memory}, a construction-time state, which tracks prior synthetic practice history and environment information. Conditioned on this state, a \emph{Proposer} generates feasible synthetic tasks that expand coverage toward under-explored aspects of the environment, and a \emph{Solver} executes these tasks to produce trajectories. A \emph{Validator} then filters out infeasible task trajectories before memory insertion, and a memory update module distills the remaining trajectories into \emph{solver memory}, the reusable procedural memory used for future tasks. In this way, \textsc{Preping} builds memory by shaping both the practice distribution and the quality of trajectories admitted into memory. 

We evaluate \textsc{Preping} on AppWorld~\citep{appworld}, BFCL v3~\citep{bfclv3}, and MCP-Universe~\citep{mcpuniverse}, covering stateful app execution, structured function calling, and realistic MCP-server tool use. The results show three main findings. First, \textsc{Preping} builds effective pre-task memory across all benchmarks, improving over a no-memory baseline by 17.1 points on AppWorld, 19.3 points on BFCL v3, and 5.4 points on MCP-Universe, while remaining competitive with methods that rely on human-defined or deployment-time target tasks, even though \textsc{Preping} requires no such target-task experience. Second, ablations show that these gains come not from synthetic task generation alone, but from validation-gated memory admission and proposer-side control over feasibility, tool coverage, and downstream relevance. Third, \textsc{Preping} offers deployment benefits as an initialization for the online memory construction approach (ACE~\cite{ace}) and as frozen pre-task memory. In particular, \textsc{Preping}+ACE improves performance on AppWorld from 71.3 to 76.3 while reducing early cold-start failures and tool-coverage shortfall (\Cref{fig:concept_fig}, Right); when frozen, \textsc{Preping} avoids deployment-time memory-update calls, reducing cost by $2.99\times$ on AppWorld and $2.23\times$ on BFCL v3 relative to ACE-Online.

\section{Related Work}

\paragraph{Memory for LLM Agents.}
Reusable memory enables LLM agents to adapt across tasks while keeping the underlying model fixed~\citep{amem,memevolve,memoryagesurvey}. As external context, memory can be inspected, revised, and transferred across models or modules, making it attractive for tool-using agents and compound AI systems~\citep{contextengineeringsurvey,ace,decomposedprompting}. Prior work stores experience in various forms, including persistent memory, workflow knowledge, playbook-style guidance, or long-term context distilled from prior interactions~\citep{memgpt,memorybank,generativeagents,storagetoexperiencesurvey}. Agent Workflow Memory induces reusable workflows from successful trajectories and retrieves them for future task solving~\citep{awm}, while ACE grows playbook-style context through structured generation, reflection, and curation from offline or online task feedback~\citep{ace}. These methods show that external memory can improve downstream execution by capturing environment-specific procedures, failure modes, and task-solving strategies~\citep{expel,clin,synapse,reflexion}. However, a key commonality is that memory is constructed from target-environment task experience, whether as curated demonstrations, logged trajectories, successful workflows, or online user interactions~\citep{memento,jefhinter,reasoningbank}. This assumption is limiting for newly connected environments, where such experience may not yet exist. In contrast, our work studies the preceding cold-start phase: constructing reusable procedural memory before any human-provided or deployment-time target tasks are available.

\paragraph{Self-Generated Practice for Policy Updates.}
A separate line of work uses self-generated tasks, self-play, and automatic curricula to improve agent policies or model behavior without human annotations~\citep{pae,absolutezero,selfevolvingagentsurvey}. In the tool-use setting, \citet{selfchallenging} instantiate this pattern with a challenger that interacts with tools to generate Code-as-Task problems with executable verification functions, and an executor that is optimized with evaluation feedback as reward. \citet{rzero} develop a related co-evolution loop for reasoning, where a Challenger is rewarded for producing tasks near a Solver's capability frontier and the Solver is trained on filtered self-generated problems. Other self-evolving systems follow related patterns in search, tool-integrated reasoning, software engineering, and corpus-grounded reasoning~\citep{drzero,agentevolver,agent0,sweselfplay,spice,toolr0}. These methods demonstrate the value of self-generated practice, but mainly as a training signal for policy or model updates, with task generation optimized for difficulty, solvability, curriculum progression, executable verification, or reward quality~\citep{nnetnav,agentgen}. In contrast, our setting requires a different form of control: since the goal is to construct reusable textual memory, synthetic practice must expose broad, non-redundant, and environment-grounded procedures, while only trajectories suitable for distillation should be admitted into memory. Therefore, our setting is not only about generating challenging or verifiable tasks, but about jointly controlling what to practice and what to store so that synthetic experience becomes deployable procedural guidance.

\section{Method}
We propose \textsc{Preping}, a framework for pre-task memory construction that turns environment access (before any target task experience) into procedural memory through controlled synthetic practice.

\subsection{Pre-Task Memory Construction}
We first formalize the pre-task memory construction setting. Given a target environment $E$ (executable) and its documentation $\mathcal{D}$, a construction procedure may inspect $\mathcal{D}$, call tools in $E$, and observe the resulting environment feedback, but it has no access to target-environment task experience, such as human-provided task instructions, demonstrations, solved trajectories, or logged user interactions. The output is a solver memory $M_{\mathrm{sol}}$ that is supplied to the agent at deployment time.

This setting differs from standard offline or online memory construction because the construction procedure has no access to the task distribution. Documentation specifies callable interfaces, but rarely reveals which tool compositions, preconditions, or failure modes will matter for downstream. The agent must therefore actively produce its own task-level objectives, execute them in the environment, and convert the resulting experience into memory. \textsc{Preping} treats this as a controlled synthetic-practice problem, where the challenge is to jointly regulate what to practice and what to store.

\begin{figure*}[t]
    \centering
    \includegraphics[width=1.0\linewidth]{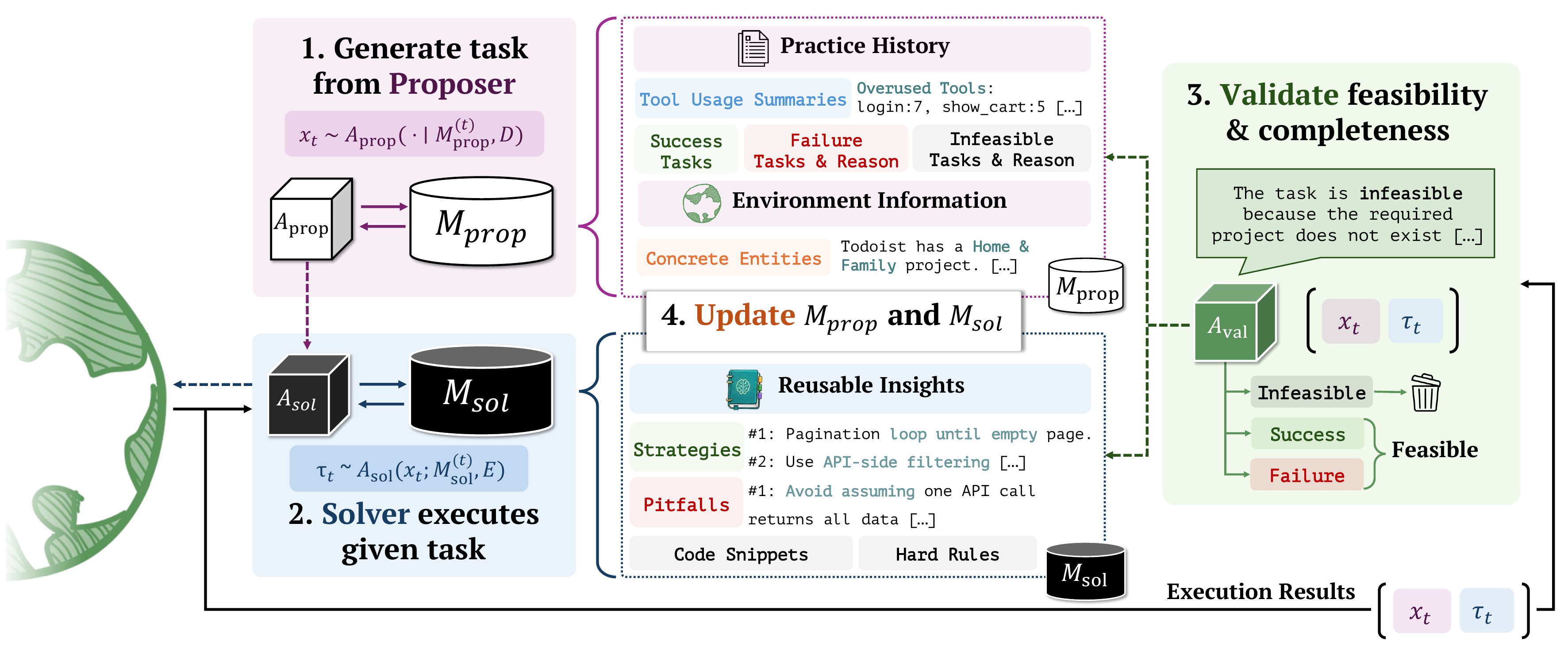}
    \vspace{-0.275in}
    \caption{
    \textbf{Overview of \textsc{Preping}.}
    \textsc{Preping} builds procedural memory before deployment through a synthetic-practice loop: proposing tasks, executing them in the environment, validating the resulting trajectories, and updating memory. Proposer memory ($M_{\mathrm{prop}}$) shapes what to practice next, while the Validator filters which synthetic trajectories are reliable enough to become solver memory ($M_{\mathrm{sol}}$).
    }
    \vspace{-0.1in}
    \label{fig:main_fig}
\end{figure*}

\subsection{\textsc{Preping}: Controlled Synthetic Practice for Pre-Task Memory Construction}

\textsc{Preping} separates the construction process into two memory states with distinct roles. Proposer memory ($M_{\mathrm{prop}}$) is a construction-time control state that guides future task proposals. It records what has already been practiced, which tools or workflows remain under-explored, and which proposals failed due to infeasibility or poor grounding. In contrast, solver memory ($M_{\mathrm{sol}}$) is the deployment-facing procedural memory that will later be provided to the task-solving agent. This separation is important because signals useful for controlling practice, such as rejected tasks, repeated tool use, and coverage imbalance, should not necessarily be exposed as procedural guidance during deployment. 

At each construction iteration, \textsc{Preping} coordinates three LLM-powered modules instantiated with different roles and contexts: a Proposer ($A_{\mathrm{prop}}$), a Solver ($A_{\mathrm{sol}}$), and a Validator ($A_{\mathrm{val}}$). At iteration $t$, the Proposer generates a synthetic task $x_t$ conditioned on documentation and proposer memory; the Solver executes $x_t$ in the environment to produce a trajectory $\tau_t$; and the Validator evaluates whether the task-trajectory pair is feasible, grounded, and useful for memory construction, as follows:

\[
x_t \sim A_{\mathrm{prop}}(\cdot \mid M_{\mathrm{prop}}^{(t)}, \mathcal{D}), \qquad
\tau_t \sim A_{\mathrm{sol}}(\cdot \mid x_t, M_{\mathrm{sol}}^{(t)}, E), \qquad
v_t = A_{\mathrm{val}}(x_t, \tau_t).
\]

The two memories are then updated asymmetrically, as follows:
\[
M_{\mathrm{prop}}^{(t+1)} = U_{\mathrm{prop}}(M_{\mathrm{prop}}^{(t)}, x_t, \tau_t, v_t), \qquad
M_{\mathrm{sol}}^{(t+1)} =
\begin{cases}
U_{\mathrm{sol}}(M_{\mathrm{sol}}^{(t)}, x_t, \tau_t, v_t), & \text{if } \mathrm{Feasible}(v_t), \\
M_{\mathrm{sol}}^{(t)}, & \text{otherwise}.
\end{cases}
\]
$U_{\mathrm{prop}}$ and $U_{\mathrm{sol}}$ denote the proposer-memory and solver-memory update rules, and $\mathrm{Feasible}(v_t)$ indicates that the synthetic task and its trajectory are grounded in the environment and suitable for memory construction. This asymmetric update is the core design of \textsc{Preping}: all experience (including rejected tasks) updates proposer memory and shapes future practice, while only feasible task-trajectory pairs are eligible for solver memory. For clarity, while equations show one synthetic task per iteration, in practice, \textsc{Preping} samples a batch of tasks as shown in \Cref{alg:full_preping}.

\subsection{Proposer Memory Controls What to Practice}

The first control decision is \emph{what to practice next}. Synthetic tasks determine which parts of the environment will be exercised and, ultimately, which procedures can be distilled into memory. If proposals repeatedly target the same tools, APIs, entities, or workflows, construction yields redundant memory with limited downstream coverage. If proposals depend on unsupported entities, unavailable tools, or hidden preconditions, they produce infeasible trajectories that provide little reusable signal. We therefore use proposer memory ($M_{\mathrm{prop}}$) as a construction-time control state to make task proposal history-aware, coverage-seeking, and grounded in the executable environment.

Proposer memory contains two complementary views of prior practice. The first is a \emph{practice-history view}, which records previous synthetic tasks, tools or APIs invoked in their trajectories, validation outcomes, and failure or infeasibility reasons. It also maintains aggregate usage summaries, identifying which tools, functions, or workflows have been over-practiced or under-practiced. Operationally, $U_{\mathrm{prop}}$ extracts invoked tools and functions from trajectories using rule-based parsers and combines them with validator feedback. The second is a \emph{grounded-environment view}, which records concrete entities, observed states, preconditions, and constraints discovered during execution. These observations are summarized with an LLM so that future proposals can refer to executable environment facts rather than inventing unsupported task details. When rendered as context for $A_{\mathrm{prop}}$, these two views impose complementary pressures: practice history discourages near-duplicate tasks and encourages expansion toward under-covered parts of the environment, while grounded environment information keeps that expansion feasible. As construction proceeds, $M_{\mathrm{prop}}$ therefore acts as a control state over the synthetic practice distribution, rather than a passive log of previous attempts.

\subsection{Validator-Gated Memory Controls What to Store}

The second control decision is \emph{what to store}. Since synthetic practice is produced without human-written task specifications or gold trajectories, its outputs are not reliable sources of memory by default. The proposed tasks may be infeasible, depend on missing environment state, require unavailable tools, or only partially specify the intended objective. If their trajectories are inserted into solver memory without filtering, synthetic artifacts can be distilled into misleading procedural guidance.

To prevent this, Validator ($A_{\mathrm{val}}$) evaluates each task-trajectory pair and produces a signal $v_t$ with feasibility and task-completion scores, along with rationales. The feasibility judgment checks whether the proposed task is grounded in the environment and executable under the observed state and available tools, while the completion judgment checks whether the Solver accomplishes the intended synthetic objective. These two judgments serve different roles: feasibility gates solver-memory insertion, while completion guides what procedural lesson, if any, should be distilled from the trajectory.

These Validator outputs are used in three ways. First, they gate solver-memory updates: infeasible pairs are excluded from $U_{\mathrm{sol}}$. Second, all validation outcomes, including rejected pairs, are passed to $U_{\mathrm{prop}}$, helping future proposals avoid repeated failure modes. Third, for admitted pairs, $U_{\mathrm{sol}}$ converts the task, trajectory, and validation outcome into compact procedural bullets (rather than appending raw interaction logs), following a reflector-curator style playbook induction process in ACE~\citep{ace}.

\section{Experiments}

\subsection{Experimental Setup}

\paragraph{Benchmarks.}
We evaluate \textsc{Preping} on three complementary agent benchmarks: AppWorld~\cite{appworld}, BFCL v3~\cite{bfclv3}, and MCP-Universe~\cite{mcpuniverse}, which span diverse forms of executable agent environments: stateful application workflows, structured function calling, and realistic MCP-server interactions. AppWorld tests stateful application tasks, where agents write code against app APIs (e.g., Spotify) and are scored by a state-based evaluator that checks the final environment state. We report AppWorld on Test-Normal (N), a held-out split drawn from the same distribution as the offline training split, and Test-Challenge (C), a harder split whose tasks require at least one unseen app. For metrics, Task Goal Completion (TGC) is the percentage of tasks for which all evaluation tests pass, while Scenario Goal Completion (SGC) credits a scenario only when all of its task variants are solved. In the main table, N-TGC/N-SGC and C-TGC/C-SGC denote TGC/SGC on Test-Normal and Test-Challenge, respectively. BFCL v3 tests executable function calling under schema and dialogue constraints; we report the Base, Long Context (Ctx.), Missing Parameter (Para.), and Missing Function (Func.) categories. MCP-Universe tests tool use over real Model Context Protocol servers with heterogeneous tool inventories and execution-based evaluators. We use four MCP-Universe categories: Repository Management (Repo.), Financial Analysis (Fin.), 3D Designing (3D.), and Browser (Brow.).

\paragraph{Pre-Task Methods.}
We compare methods that operate strictly within the pre-task setting, where no target-environment task data are available. \textbf{Base} uses no constructed memory and solves downstream tasks directly from the model and environment context. \textbf{Direct Memory} constructs memory directly from environment documentation without execution, by sampling and combining diverse subsets of API or tool documentation into memory. We also evaluate execution-based baselines that construct memory from free-form environment interaction without task-level objectives. Specifically, \textbf{Random Exploration} prompts the agent to explore the environment without additional constraints, while \textbf{Guided Exploration} conditions exploration on prior exploration history to encourage under-explored APIs or tools. \textbf{\textsc{Preping}} instead constructs memory from proposer-guided synthetic task practice: it generates task-level objectives, executes them in the environment, and admits only validator-approved task-trajectory pairs into solver memory. All memory-construction methods use the same reflector-curator memory induction pipeline~\citep{ace}, isolating whether memory is induced from documentation, free-form exploration, or validated synthetic-task practice. We provide details in \Cref{sec:baseline_details}.

\paragraph{Task-Informed Methods.}
We also report task-informed memory construction methods as reference points. Unlike the pre-task methods above, these methods are allowed to use target-environment task data, and therefore assume information that is unavailable in the pre-task setting. \textbf{ACE-Offline}~\citep{ace} constructs memory before deployment from human-defined target tasks and their execution feedback. This setting can produce strong memory when a representative task set is available, but it requires task collection or task design for each new environment. In our benchmark suite, we evaluate it only on AppWorld, the only benchmark that provides a training split. \textbf{ACE-Online}~\citep{ace} constructs memory during deployment from user tasks as they arrive. This removes the need for a pre-collected task set, but adds deployment-time memory-construction cost and latency, and the agent begins with empty memory, exposing early failures to users during the cold-start period.

\begin{table}[t!]
    \centering
    \caption{\textbf{Main results.} Pre-Task Methods build memory without task data, while Task-Informed Methods use task data before or during deployment. Scores are averaged over three independent runs. Bold denotes the best pre-task method.}
    \label{tab:main_result}
    \setlength{\tabcolsep}{3.0pt}
    \begingroup
    \renewcommand{\arraystretch}{1.25}
    \resizebox{\linewidth}{!}{%
    \begin{tabular}{l*{4}{c}>{\columncolor{avggray}}c*{4}{c}>{\columncolor{avggray}}c*{4}{c}>{\columncolor{avggray}}c}
        \toprule
        & \multicolumn{5}{c}{\textbf{AppWorld}} & \multicolumn{5}{c}{\textbf{BFCL v3}} & \multicolumn{5}{c}{\textbf{MCP-Universe}} \\
        \cmidrule(l{2pt}r{2pt}){2-6}
        \cmidrule(l{2pt}r{2pt}){7-11}
        \cmidrule(l{2pt}r{2pt}){12-16}
        \textbf{Method}
        & \textbf{N-TGC}
        & \textbf{N-SGC}
        & \textbf{C-TGC}
        & \textbf{C-SGC}
        & \textbf{Avg.}
        & \textbf{Base}
        & \textbf{Ctx.}
        & \textbf{Para.}
        & \textbf{Func.}
        & \textbf{Avg.}
        & \textbf{Repo.}
        & \textbf{Fin.}
        & \textbf{3D.}
        & \textbf{Brow.}
        & \textbf{Avg.} \\
        \midrule
        \rowblue \multicolumn{16}{c}{\textit{\textbf{Pre-Task Methods}}} \\
        Base & 69.6 & 49.4 & 56.7 & 36.7 & 53.1 & 43.7 & 40.0 & 23.0 & 28.3 & 33.8 & 8.1 & 59.2 & 29.8 & 31.3 & 32.1 \\
        Direct Memory & 72.8 & 55.4 & 56.0 & 34.5 & 54.7 & 43.2 & 42.3 & 22.3 & 29.3 & 34.3 & 7.1 & 67.5 & 26.3 & 30.2 & 32.8 \\
        Random Exploration & 73.6 & 52.4 & 55.9 & 36.9 & 54.7 & 59.0 & 51.3 & 32.8 & 41.3 & 46.1 & 10.1 & 60.8 & 22.8 & \textbf{38.5} & 33.1 \\
        Guided Exploration & 74.6 & 56.0 & 56.6 & 35.0 & 55.6 & 55.5 & 51.2 & 28.5 & 40.3 & 43.9 & \textbf{12.1} & 61.7 & 33.3 & 31.3 & 34.6 \\
        \textbf{\textsc{Preping}} & \textbf{83.7} & \textbf{70.2} & \textbf{72.2} & \textbf{54.7} & \textbf{70.2} & \textbf{65.2} & \textbf{59.3} & \textbf{37.8} & \textbf{50.2} & \textbf{53.1} & 10.1 & \textbf{70.8} & \textbf{36.8} & 32.3 & \textbf{37.5} \\
        \midrule
        \rowpink \multicolumn{16}{c}{\textit{\textbf{Task-Informed Methods}}} \\
        ACE-Offline & 82.7 & 69.1 & 69.0 & 50.3 & 67.8 & -- & -- & -- & -- & -- & -- & -- & -- & -- & -- \\
        ACE-Online & 80.4 & 65.5 & 78.3 & 60.9 & 71.3 & 62.3 & 55.0 & 36.7 & 52.3 & 51.6 & 14.1 & 69.2 & 43.9 & 37.5 & 41.2 \\
        \bottomrule
    \end{tabular}
    }
    \endgroup
    \vspace{-0.05in}
\end{table}

\paragraph{Implementation Details.}
We use DeepSeek-V3.2~\cite{deepseekv32} without reasoning mode as the base LLM for all components, including the Proposer, Solver, Validator, and memory-update calls in \textsc{Preping}. The same model is also used across all methods for task execution and (when applicable) memory construction. The main results are averaged over three independent construction-and-evaluation runs. We use temperature 0 for Solver execution, 0.7 for Validator judgments and memory updates, and 1.0 for synthetic task proposal. The Validator assigns 1-5 Likert scores for task feasibility and task completion. Trajectories are admitted into solver memory at feasibility scores of 5 for AppWorld and MCP-Universe and at least 2 for BFCL v3, while completion scores of at least 4 indicate successful execution. Each \textsc{Preping} construction run uses 10 iterations with 10 synthetic tasks per iteration, yielding 100 synthetic practice tasks in total. To preserve the pre-task setting, pre-task methods may use only environment documentation $\mathcal{D}$, such as tool descriptions or API documentation, and feedback obtained from self-generated interactions during memory construction. Additional implementation details, including the full construction algorithm and prompt templates, are provided in \Cref{sec:additional_experimental_details}.

\subsection{Experimental Results and Analyses}

\paragraph{\textsc{Preping} Builds Strong Memory Without Target-Task Experience.}
\Cref{tab:main_result} shows that pre-task methods based on documentation or free-form exploration provide limited gains. Direct Memory converts documentation into memory without observing tool execution, which limits its ability to capture procedural knowledge about tool behavior, state change, and failure conditions. Random and Guided Exploration benefit from execution feedback (especially on BFCL v3), but their interaction is not organized around task-level objectives, often leading to shallow, repetitive, or weakly compositional tool use (rather than reusable workflows involving multi-step tool composition). In contrast, \textsc{Preping} constructs memory from proposer-guided synthetic tasks and validation-filtered trajectories, achieving the strongest pre-task performance across all three benchmarks and improving average score over Base by 17.1 points on AppWorld, 19.3 points on BFCL v3, and 5.4 points on MCP-Universe. Furthermore, despite using no human-defined or deployment-time user tasks for memory construction, \textsc{Preping} remains competitive with task-informed methods. Specifically, on AppWorld, \textsc{Preping} exceeds ACE-Offline and is close to ACE-Online, while on BFCL v3, it surpasses ACE-Online on average. These results suggest that \textsc{Preping} can produce reusable procedural memory that supports downstream task solving even before any target-task experience is available.

\begin{table}[t]
    \centering
    \caption{\textbf{Component ablation.} Score uses AppWorld Test-Normal TGC/SGC and BFCL v3 Base success rate. Env. Info denotes environment information. \protect\smallhl{cappink}{Naive Task Generation} disables all components; \protect\smallhl{capblue}{\textsc{Preping}} enables all. Bold and underline mark the best and second-best values.}
    \label{tab:component_ablation}
    \setlength{\tabcolsep}{3.0pt}
    \begingroup
    \newcommand{\yescomp}{\textcolor{Green}{\textbf{\ding{51}}}}
    \newcommand{\nocomp}{\textcolor{Red}{\textbf{\ding{55}}}}
    \newcommand{\naivecell}[1]{\cellcolor{mypink}#1}
    \newcommand{\prepingcell}[1]{\cellcolor{myblue}#1}
    \renewcommand{\arraystretch}{1.25}
    \resizebox{\linewidth}{!}{%
    \begin{tabular}{lcccccccc}
        \toprule
        \textbf{Benchmark} & \textbf{Validator} & \textbf{Env. Info} & \textbf{Practice History} & \textbf{Score $\uparrow$} & \textbf{Infeasible Task (\%) $\downarrow$} & \textbf{Unique Tools $\uparrow$} & \textbf{Tool Entropy $\uparrow$} & \textbf{W. Recall $\uparrow$} \\
        \midrule
        \multirow{5}{*}{AppWorld}
        & \naivecell{\nocomp} & \naivecell{\nocomp} & \naivecell{\nocomp} & \naivecell{47.8 / 26.8} & \naivecell{--} & \naivecell{65.7} & \naivecell{5.611} & \naivecell{0.653} \\
        & \yescomp & \nocomp & \nocomp & 78.2 / 60.7 & 26.3 & 69.0 & 5.586 & 0.632 \\
        & \yescomp & \nocomp & \yescomp & \underline{81.5} / \underline{67.9} & 33.3 & \underline{81.7} & \underline{5.775} & \underline{0.674} \\
        & \yescomp & \yescomp & \nocomp & 80.3 / 64.9 & \textbf{5.3} & 42.3 & 4.711 & 0.368 \\
        & \prepingcell{\yescomp} & \prepingcell{\yescomp} & \prepingcell{\yescomp} & \prepingcell{\textbf{83.7 / 70.2}} & \prepingcell{\underline{16.0}} & \prepingcell{\textbf{87.0}} & \prepingcell{\textbf{5.919}} & \prepingcell{\textbf{0.703}} \\
        \midrule
        \multirow{5}{*}{BFCL v3}
        & \naivecell{\nocomp} & \naivecell{\nocomp} & \naivecell{\nocomp} & \naivecell{59.5} & \naivecell{--} & \naivecell{47.0} & \naivecell{5.022} & \naivecell{0.526} \\
        & \yescomp & \nocomp & \nocomp & 62.0 & 9.0 & 44.7 & 5.015 & 0.505 \\
        & \yescomp & \nocomp & \yescomp & 60.8 & 19.0 & \textbf{118.0} & \underline{6.028} & \textbf{0.968} \\
        & \yescomp & \yescomp & \nocomp & \underline{64.0} & \underline{6.0} & 50.3 & 5.022 & 0.591 \\
        & \prepingcell{\yescomp} & \prepingcell{\yescomp} & \prepingcell{\yescomp} & \prepingcell{\textbf{65.2}} & \prepingcell{\textbf{4.0}} & \prepingcell{\underline{105.0}} & \prepingcell{\textbf{6.095}} & \prepingcell{\underline{0.846}} \\
        \bottomrule
    \end{tabular}
    }
    \endgroup
    \vspace{-0.05in}
\end{table}

\begin{figure}[t]
    \centering

    \begin{minipage}[t]{0.58\linewidth}
        \vspace{0pt}
        \captionof{table}{\textbf{\textsc{Preping} warm-starts online memory construction.} \textsc{Preping}+ACE initializes ACE-Online with pre-task memory and applies the standard online update procedure.}
        \label{tab:preping_online}
        \centering
\setlength{\tabcolsep}{3.0pt}
\begingroup
\renewcommand{\arraystretch}{1.2}
\resizebox{\linewidth}{!}{%
\begin{tabular}{l*{4}{c}>{\columncolor{avggray}}c*{2}{c}>{\columncolor{avggray}}c}
    \toprule
    & \multicolumn{5}{c}{\textbf{AppWorld}} & \multicolumn{3}{c}{\textbf{BFCL v3}} \\
    \cmidrule(l{2pt}r{2pt}){2-6}
    \cmidrule(l{2pt}r{2pt}){7-9}
    \textbf{Method}
    & N-TGC
    & N-SGC
    & C-TGC
    & C-SGC
    & Avg.
    & Base
    & Ctx.
    & Avg. \\
    \midrule
    \rowblue \multicolumn{9}{c}{\textit{\textbf{Pre-Task Methods}}} \\
    Base & 69.6 & 49.4 & 56.7 & 36.7 & 53.1 & 43.7 & 40.0 & 41.9 \\
    \textsc{Preping} & 83.7 & 70.2 & 72.2 & 54.7 & 70.2 & 65.2 & 59.3 & 62.3 \\
    \midrule
    \rowpink \multicolumn{9}{c}{\textit{\textbf{Task-Informed Methods}}} \\
    ACE-Online & 80.4 & 65.5 & 78.3 & 60.9 & 71.3 & 62.3 & 55.0 & 58.7 \\
    \textbf{\textsc{Preping}+ACE} & \textbf{86.1} & \textbf{73.8} & \textbf{80.1} & \textbf{65.2} & \textbf{76.3} & \textbf{66.0} & \textbf{63.7} & \textbf{64.9} \\
    \bottomrule
\end{tabular}
}
\endgroup
\vspace{-0.1in}

    \end{minipage}
    \hfill
    \begin{minipage}[t]{0.4\linewidth}
        \vspace{0pt}
        \centering
        \vspace{0.1in}
        \includegraphics[width=0.99\linewidth]{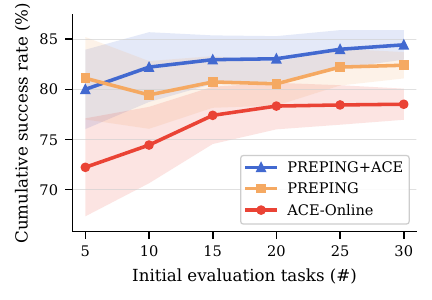}
        \vspace{-0.28in}
        \captionof{figure}{\textbf{\textsc{Preping} mitigates online cold-start before memory builds up.}}
        \label{fig:appworld_coldstart30_prefix_curve}
        \vspace{-0.08in}
    \end{minipage}

\end{figure}
\paragraph{Ablation Study: Controlled and Validated Practice Drives the Gains.}

We next ask which parts of \textsc{Preping} are responsible for the gains. \Cref{tab:component_ablation} ablates three components: validation-gated memory admission, practice history, and environment information. The all-disabled rows correspond to Naive Task Generation, while the full rows enable all three components. To analyze deeply beyond final task performance, we also report four construction-side diagnostics: infeasible task rate, unique tools, tool entropy, and weighted recall over tools used in deployment-time test trajectories. Infeasible task rate measures how often the validator rejects generated tasks as infeasible. Unique tools counts distinct AppWorld domain APIs or BFCL functions invoked in synthetic trajectories, and tool entropy measures how evenly practice is distributed across them. Weighted recall (W. Recall) compares the tools covered during construction with those used in deployment-time test trajectories, weighting each covered tool by its test-time frequency. Together, these metrics separate three desiderata for pre-task memory construction: task quality, support expansion, and relevance to downstream workflows.

\Cref{tab:component_ablation} first shows that validation-gated admission is crucial for constructing reusable memory. On AppWorld, adding Validator to Naive Task Generation improves performance from 47.8/26.8 to 78.2/60.7. This gain cannot be explained by tool breadth alone: Naive already invokes 65.7 unique APIs with a weighted recall of 0.653, but its unfiltered trajectories can distill infeasible objectives, constraint violations, or failed recovery patterns into solver memory. \Cref{sec:no_validator_qualitative_example} illustrates this failure mode. On BFCL v3, validation also improves average performance from 59.5 to 62.0, although the gain is smaller because infeasible task rates are lower in this setting.

Beyond validation, proposer memory determines which parts of the environment are practiced. Practice history primarily expands support by discouraging repeated practice and pushing proposals toward under-covered tools, increasing AppWorld unique APIs from 69.0 to 81.7 and BFCL v3 from 44.7 to 118.0. Yet, history alone can over-expand into poorly grounded proposals because it tracks what has been practiced but not necessarily what is executable in the current environment. Environmental information has a complementary effect: it keeps proposals anchored to observed entities, states, and constraints, but by itself lacks the history needed to broaden practice. \textsc{Preping} combines both signals, yielding the best downstream performance on both benchmarks while maintaining strong tool entropy and weighted recall. These results show that effective pre-task memory construction requires more than generating synthetic tasks at scale; it requires balancing feasibility, coverage expansion, and relevance to workflows the solver will later encounter. We provide iteration-wise construction dynamics for these ablations in \Cref{sec:component_ablation_iteration_dynamics}, further decompose the Validator's role in \Cref{sec:validator_signal_ablation}, and present an AppWorld example of this proposer-memory interaction in \Cref{sec:appworld_proposer_memory_qualitative_example}.

\paragraph{\textsc{Preping}+ACE Improves Online Learning Through Better Initialization.}
\textsc{Preping} can also serve as an initialization for online memory construction. Before deployment, it constructs reusable solver memory from synthetic practice without target-environment tasks. During deployment, this memory can then continue to be updated as real user tasks arrive. We call this setting \textsc{Preping}+ACE, initializing ACE-Online with \textsc{Preping} memory and then running the standard online update procedure during evaluation. As shown in \Cref{tab:preping_online}, \textsc{Preping}+ACE improves AppWorld average performance from 71.3 to 76.3. The improvement is especially clear on AppWorld Test-Challenge, where it surpasses ACE-Online on both TGC and SGC, increasing TGC from 78.3 to 80.1 and SGC from 60.9 to 65.2. On the evaluated BFCL v3 categories, the same initialization improves Base performance from 62.3 to 66.0 and Long Context performance from 55.0 to 63.7, raising their average from 58.7 to 64.9. These results show that \textsc{Preping} is not only useful as a frozen deployment artifact, but also as a strong starting point for online memory construction.

\begin{figure}[t!]
    \centering
    \begin{minipage}[t]{0.56\linewidth}
        \captionof{table}{\textbf{\textsc{Preping} generalizes across backbone models.} Results are reported on AppWorld Test-Normal with each method run using a single backbone model.}
        \vspace{-0.1in}
        \label{tab:appworld_other_models}
        \centering
\setlength{\tabcolsep}{3pt}
\newlength{\backboneGptMetricWidth}
\newlength{\backboneOssMetricWidth}
\newlength{\backboneQwenMetricWidth}
\settowidth{\backboneGptMetricWidth}{\textbf{57.7}}
\settowidth{\backboneOssMetricWidth}{\textbf{GPT-OSS-120B}}
\settowidth{\backboneQwenMetricWidth}{\textbf{Qwen3-235B-A22B}}
\setlength{\backboneOssMetricWidth}{\dimexpr\backboneOssMetricWidth/2-\tabcolsep\relax}
\setlength{\backboneQwenMetricWidth}{\dimexpr\backboneQwenMetricWidth/2-\tabcolsep\relax}
\resizebox{\linewidth}{!}{%
\begin{tabular}{l*{2}{w{c}{\backboneGptMetricWidth}}*{2}{w{c}{\backboneOssMetricWidth}}*{2}{w{c}{\backboneQwenMetricWidth}}}
    \toprule
    & \multicolumn{2}{c}{\textbf{GPT-5.1}} & \multicolumn{2}{c}{\textbf{GPT-OSS-120B}} & \multicolumn{2}{c}{\textbf{Qwen3-235B-A22B}} \\
    \cmidrule(l{2pt}r{2pt}){2-3}
    \cmidrule(l{2pt}r{2pt}){4-5}
    \cmidrule(l{2pt}r{2pt}){6-7}
    \textbf{Method} & TGC & SGC & TGC & SGC & TGC & SGC \\
    \midrule
    \rowblue \multicolumn{7}{c}{\textit{\textbf{Pre-Task Methods}}} \\
    Base & 52.4 & 30.4 & 28.0 & 7.1 & 58.3 & 37.5 \\
    \textbf{\textsc{Preping}} & \textbf{57.7} & \textbf{41.1} & \textbf{36.9} & \textbf{17.9} & \textbf{67.3} & \textbf{44.6} \\
    \midrule
    \rowpink \multicolumn{7}{c}{\textit{\textbf{Task-Informed Methods}}} \\
    ACE-Offline & 56.0 & 37.5 & 36.3 & 16.1 & 69.6 & 50.0 \\
    ACE-Online & 52.4 & 33.9 & 31.5 & 12.5 & 66.1 & 44.6 \\
    \bottomrule
\end{tabular}
}
\vspace{-0.1in}

    \end{minipage}
    \hfill
    \begin{minipage}[t]{0.38\linewidth}
        \vspace{0pt}
        \centering
        \includegraphics[width=0.9\linewidth]{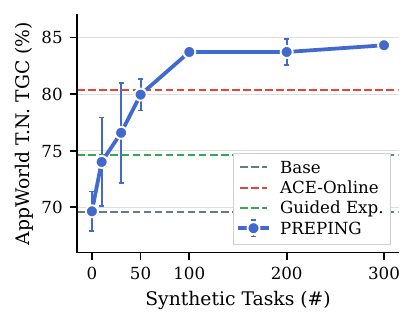}
        \vspace{-0.2in}
        \captionof{figure}{\textbf{Construction budget.} AppWorld Test-Normal TGC as a function of the number of synthetic tasks.}
        \label{fig:preping_learning_dynamics}
    \end{minipage}
    \vspace{-0.2in}
\end{figure}

\paragraph{\textsc{Preping} Reduces Early Online Cold-Start Failures.}
\Cref{fig:appworld_coldstart30_prefix_curve} examines the early deployment regime, before ACE-Online has accumulated sufficient task-informed memory. We construct 18 shuffled 30-task streams from AppWorld Test-Normal and measure cumulative success over the first $k$ tasks in each stream. Then, ACE-Online shows a clear cold start: its first 10-task success rate is 74.4\%, much closer to the no-memory Base at 69.6\% than to its full Test-Normal TGC of 80.4\%. In contrast, \textsc{Preping} starts with pre-task memory, and \textsc{Preping}+ACE further updates it online, achieving 79.4\% and 82.2\% over the first 10 tasks, respectively. This advantage persists through the full 30-task stream, showing that \textsc{Preping} and \textsc{Preping}+ACE mitigate online cold start failures by giving the agent useful procedural memory before substantial user-task experience is available. 

\paragraph{\textsc{Preping} Reduces Tool-Coverage Cold Start.}
We next examine whether early performance gaps reflect limited tool coverage in online memory. The right panel of \Cref{fig:coverage_cold_start} shows that ACE-Online starts with empty memory and only accumulates coverage from evaluation tasks, while \textsc{Preping} enters deployment with coverage from 100 synthetic practice tasks, and \textsc{Preping}+ACE starts from this and further expands it with user-task updates. On AppWorld Test-Normal, ACE-Online requires 58 evaluation tasks to match \textsc{Preping}'s pre-deployment coverage; on BFCL v3 Base, it still falls short after all 200 tasks. This shows that synthetic practice reduces the coverage lag of online memory, and that \textsc{Preping}+ACE benefits by combining broad pre-task coverage with task-informed evidence.

\paragraph{Generalization Across Backbone Models.}
We further test whether \textsc{Preping} transfers across backbone models rather than relying on a single primary LLM. On AppWorld Test-Normal, we run the same, full construction-and-evaluation pipeline with GPT-5.1~\cite{gpt51}, GPT-OSS-120B~\cite{gptoss120b}, and Qwen3-235B-A22B~\cite{qwen3}, with reasoning-disabled modes for GPT-5.1 and Qwen3-235B-A22B. For each run, the same backbone instantiates the Proposer, Solver, Validator, and memory-update calls, matching the single-model setup used in our main experiments. As shown in \Cref{tab:appworld_other_models}, \textsc{Preping} improves over the no-memory Base across all three backbones and remains comparable to task-informed methods, despite using no human-defined tasks during construction. The improvement also holds for GPT-OSS-120B, the relatively weaker backbone in this comparison, indicating that \textsc{Preping} can benefit agents even when the base solver is less capable. Overall, these results suggest that the benefit comes from the controlled memory-construction procedure rather than a model-specific artifact.

\paragraph{Modest Synthetic-Task Budgets Already Yield Strong Gains.}
To understand how many synthetic tasks are needed to construct effective memory, we vary the construction budget by the number of synthetic tasks. As shown in \Cref{fig:preping_learning_dynamics}, \textsc{Preping} improves steadily as the construction progresses, reaching 76.6 after only 30 synthetic tasks and already surpassing Guided Exploration. With 50 synthetic tasks, it reaches 80.0, approaching the ACE-Online reference at 80.6. This suggests that proposer-guided practice can expose broad and useful procedures even under modest construction budgets. Additional practice continues to improve performance, with 300 synthetic tasks reaching 84.3, albeit with smaller marginal gains. Thus, the 100-task budget used in our main experiments provides a practical balance between construction cost and performance gain.

\begin{wraptable}{r}{0.36\textwidth}
    \centering
    \captionsetup{justification=raggedright,singlelinecheck=false}
    \centering
\vspace{-0.1in}

\captionof{table}{\textbf{\textsc{Preping} with a 10 offline task set.} Results are averaged over three runs.}
\label{tab:preping_offline_init}
\vspace{-0.1in}
\setlength{\tabcolsep}{4pt}

\begin{tabular}{lcc}
    \toprule
    \textbf{Method} & \textbf{TGC} & \textbf{SGC} \\
    \midrule
    Base & 69.6 & 49.4 \\
    \textsc{Preping} & 83.7 & 70.2 \\
    Task-Seeded \textsc{Preping} & \textbf{85.1} & \textbf{73.8} \\
    \bottomrule
\end{tabular}

\end{wraptable}

\paragraph{Task-Seeded \textsc{Preping} Guides Synthetic Practice.}
Although constructing a large, diverse offline task set can be costly, a small seed set may be available before deployment. We therefore evaluate Task-Seeded \textsc{Preping}, which starts from 10 solved tasks (sampled from the AppWorld training split) and then runs the same iterative \textsc{Preping} loop as in the main setting. It initializes proposer memory from the sampled task instructions and solver memory from their corresponding trajectories. As shown in \Cref{tab:preping_offline_init}, Task-Seeded \textsc{Preping} improves AppWorld Test-Normal TGC/SGC from 83.7/70.2 to 85.1/73.8. The result suggests that \textsc{Preping} can leverage a small offline set as an initial anchor for synthetic task proposal, while still expanding solver memory through subsequent synthetic practice.

\Needspace{17\baselineskip}
\begin{wrapfigure}{r}{0.3\textwidth}
    \vspace{-0.23in}
    \centering
    \captionsetup{justification=raggedright,singlelinecheck=false}
    \includegraphics[width=\linewidth]{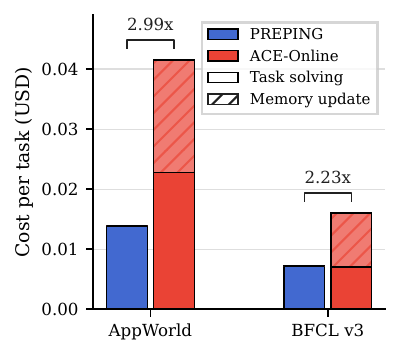}
    \vspace{-0.3in}
    \caption{\textbf{Deployment time cost per task.}}
    \vspace{-0.3in}
    \label{fig:cost_bar_plot}
\end{wrapfigure}
\paragraph{\textsc{Preping} Reduces Deployment-Time Cost.}
Online memory construction can improve performance, but it also adds memory-update calls during user-facing deployment. In contrast, \textsc{Preping} constructs solver memory before deployment, allowing user tasks to be executed without additional memory-construction calls. As shown in \Cref{fig:cost_bar_plot}, this substantially reduces deployment-time cost per task, with ACE-Online costing about $2.99\times$ more on AppWorld and $2.23\times$ more on BFCL v3. This comparison isolates deployment-time cost from the one-time pre-deployment construction cost of \textsc{Preping}. Even when this construction cost is included, \textsc{Preping} remains cheaper than ACE-Online on both AppWorld and BFCL v3, as shown in \Cref{tab:cost_construction_eval}. Thus, \textsc{Preping} shifts memory construction from a recurring deployment-time expense to an amortizable pre-deployment investment.

\section{Conclusion}

We studied \emph{pre-task memory construction}, the problem of building reusable agent memory before any target-environment task data are available. \textsc{Preping} addresses this setting by treating memory construction as a control problem over what to practice and what to store: proposer memory shapes synthetic practice, the solver executes it in the target environment, and the validator filters out infeasible trajectories before solver-memory updates. Experiments across AppWorld, BFCL v3, and MCP-Universe show that procedural memory can be constructed before deployment, outperforming pre-task baselines and providing a strong initialization for online memory adaptation. These results suggest that agent memory need not be accumulated only passively from deployment-time interactions; it can also be actively prepared through controlled, environment-grounded practice before deployment. We discuss limitations and broader impact in \Cref{sec:limitations}.

\clearpage
\section*{Acknowledgements}
This work was supported by Institute for Information \& Communications Technology Planning \& Evaluation (IITP) grants funded by the Korea government (MSIT) (RS-2019-II190075, Artificial Intelligence Graduate School Program (KAIST); RS-2022-II220713, Meta-learning Applicable to Real-world Problems; RS-2024-00469482 \& RS-2024-00509279, Global AI Frontier Lab International Collaborative Research), and by National Research Foundation of Korea (NRF) grants funded by the Korea government (MSIT) (RS-2023-00256259; RS-2026-25488933). Support was also provided by the InnoCORE program of the Ministry of Science and ICT (No. N10250156) and the ``Advanced GPU Utilization Support Program'' funded by the Government of the Republic of Korea (Ministry of Science and ICT).
\bibliography{reference}

@string{neurips = {Neural Information Processing Systems (NeurIPS)}}

@string{icml    = {International Conference on Machine Learning (ICML)}}

@string{iclr    = {International Conference on Learning Representations (ICLR)}}

@string{acl     = {Annual Meeting of the Association for Computational Linguistics (ACL)}}

@string{arxiv   = {arXiv}}

@article{rzero,
  title={R-Zero: Self-Evolving Reasoning {LLM} from Zero Data},
  author={Huang, Chengsong and Yu, Wenhao and Wang, Xiaoyang and Zhang, Hongming and Li, Zongxia and Li, Ruosen and Huang, Jiaxin and Mi, Haitao and Yu, Dong},
  journal=arxiv,
  year={2025},
  doi={10.48550/arXiv.2508.05004},
}

@article{drzero,
  title={Dr. Zero: Self-Evolving Search Agents without Training Data},
  author={Yue, Zhenrui and Upasani, Kartikeya and Yang, Xianjun and Ge, Suyu and Nie, Shaoliang and Mao, Yuning and Liu, Zhe and Wang, Dong},
  journal=arxiv,
  year={2026},
  doi={10.48550/arXiv.2601.07055},
}

@inproceedings{reasoningbank,
  title={ReasoningBank: Scaling Agent Self-Evolving with Reasoning Memory},
  author={Ouyang, Siru and Yan, Jun and Hsu, I-Hung and Chen, Yanfei and Jiang, Ke and Wang, Zifeng and Han, Rujun and Le, Long T. and Daruki, Samira and Tang, Xiangru and Tirumalashetty, Vishy and Lee, George and Rofouei, Mahsan and Lin, Hangfei and Han, Jiawei and Lee, Chen-Yu and Pfister, Tomas},
  booktitle=iclr,
  year={2026},
}

@article{agentevolver,
  title={AgentEvolver: Towards Efficient Self-Evolving Agent System},
  author={Zhai, Yunpeng and Tao, Shuchang and Chen, Cheng and Zou, Anni and Chen, Ziqian and Fu, Qingxu and Mai, Shinji and Yu, Li and Deng, Jiaji and Cao, Zouying and Liu, Zhaoyang and Ding, Bolin and Zhou, Jingren},
  journal=arxiv,
  year={2025},
  doi={10.48550/arXiv.2511.10395},
}

@inproceedings{selfchallenging,
  title={Self-Challenging Language Model Agents},
  author={Zhou, Yifei and Levine, Sergey and Weston, Jason E. and Li, Xian and Sukhbaatar, Sainbayar},
  booktitle=neurips,
  year={2025},
}

@article{memevolve,
  title={MemEvolve: Meta-Evolution of Agent Memory Systems},
  author={Zhang, Guibin and Ren, Haotian and Zhan, Chong and Zhou, Zhenhong and Wang, Junhao and Zhu, He and Zhou, Wangchunshu and Yan, Shuicheng},
  journal=arxiv,
  year={2025},
  doi={10.48550/arXiv.2512.18746},
}

@article{agent0,
  title={Agent0: Unleashing Self-Evolving Agents from Zero Data via Tool-Integrated Reasoning},
  author={Xia, Peng and Zeng, Kaide and Liu, Jiaqi and Qin, Can and Wu, Fang and Zhou, Yiyang and Xiong, Caiming and Yao, Huaxiu},
  journal=arxiv,
  year={2025},
  doi={10.48550/arXiv.2511.16043},
}

@article{sweselfplay,
  title={Toward Training Superintelligent Software Agents through Self-Play {SWE-RL}},
  author={Yuxiang Wei and
    Zhiqing Sun and
    Emily McMilin and
    Jonas Gehring and
    David Zhang and
    Gabriel Synnaeve and
    Daniel Fried and
    Lingming Zhang and
    Sida I. Wang},
  journal=arxiv,
  year={2025},
  doi={10.48550/arXiv.2512.18552},
}

@article{toolr0,
  title={{Tool-R0}: Self-Evolving {LLM} Agents for Tool-Learning from Zero Data},
  author={Acikgoz, Emre Can and Qian, Cheng and H{\"u}botter, Jonas and Ji, Heng and Hakkani-T{\"u}r, Dilek and Tur, Gokhan},
  journal=arxiv,
  year={2026},
  doi={10.48550/arXiv.2602.21320},
}

@inproceedings{ace,
  title={Agentic Context Engineering: Evolving Contexts for Self-Improving Language Models},
  author={Zhang, Qizheng and Hu, Changran and Upasani, Shubhangi and Ma, Boyuan and Hong, Fenglu and Kamanuru, Vamsidhar and Rainton, Jay and Wu, Chen and Ji, Mengmeng and Li, Hanchen and Thakker, Urmish and Zou, James and Olukotun, Kunle},
  booktitle=iclr,
  year={2026},
}

@article{dynamiccheatsheet,
  title={Dynamic Cheatsheet: Test-Time Learning with Adaptive Memory},
  author={Suzgun, Mirac and Y{\"u}ksekg{\"o}nul, Mert and Bianchi, Federico and Jurafsky, Dan and Zou, James},
  journal=arxiv,
  year={2025},
  doi={10.48550/arXiv.2504.07952},
}

@article{memento,
  title={Memento: Fine-tuning {LLM} Agents without Fine-tuning {LLM}s},
  author={Zhou, Huichi and Chen, Yihang and Guo, Siyuan and Yan, Xue and Lee, Kin Hei and Wang, Zihan and Lee, Ka Yiu and Zhang, Guchun and Shao, Kun and Yang, Linyi and Wang, Jun},
  journal=arxiv,
  year={2025},
  doi={10.48550/arXiv.2508.16153},
}

@article{jefhinter,
  title={{JEF}-Hinter: Leveraging Offline Knowledge for Improving Web Agents Adaptation},
  author={Nekoei, Hadi and Jaiswal, Aman and B{\'e}chard, Patrice and Shliazhko, Oleh and Ayala, Orlando Marquez and Reymond, Mathieu and Caccia, Massimo and Drouin, Alexandre and Chandar, Sarath and Lacoste, Alexandre},
  journal=arxiv,
  year={2025},
  doi={10.48550/arXiv.2510.04373},
}

@article{contextengineeringsurvey,
  title={A Survey of Context Engineering for Large Language Models},
  author={Mei, Lingrui and Yao, Jiayu and Ge, Yuyao and Wang, Yiwei and Bi, Baolong and Cai, Yujun and Liu, Jiazhi and Li, Mingyu and Li, Zhong-Zhi and Zhang, Duzhen and Zhou, Chenlin and Mao, Jiayi and Xia, Tianze and Guo, Jiafeng and Liu, Shenghua},
  journal=arxiv,
  year={2025},
  doi={10.48550/arXiv.2507.13334},
}

@inproceedings{decomposedprompting,
  title={Decomposed Prompting: A Modular Approach for Solving Complex Tasks},
  author={Khot, Tushar and Trivedi, Harsh and Finlayson, Matthew and Fu, Yao and Richardson, Kyle and Clark, Peter and Sabharwal, Ashish},
  booktitle=iclr,
  year={2023},
}

@article{gmemory,
  title={G-Memory: Tracing Hierarchical Memory for Multi-Agent Systems},
  author={Zhang, Guibin and Fu, Muxin and Wan, Guancheng and Yu, Miao and Wang, Kun and Yan, Shuicheng},
  journal=arxiv,
  year={2025},
  doi={10.48550/arXiv.2506.07398},
}

@article{legomem,
  title={LEGOMem: Modular Procedural Memory for Multi-agent {LLM} Systems for Workflow Automation},
  author={Han, Dongge and Couturier, Camille and Madrigal D{\'i}az, Daniel and Zhang, Xuchao and R{\"u}hle, Victor and Rajmohan, Saravan},
  journal=arxiv,
  year={2025},
  doi={10.48550/arXiv.2510.04851},
}

@article{memgpt,
  title={{MemGPT}: Towards {LLM}s as Operating Systems},
  author={Packer, Charles and Wooders, Sarah and Lin, Kevin and Fang, Vivian and Patil, Shishir G. and Stoica, Ion and Gonzalez, Joseph E.},
  journal=arxiv,
  year={2023},
  doi={10.48550/arXiv.2310.08560},
}

@inproceedings{memorybank,
  title={{MemoryBank}: Enhancing Large Language Models with Long-Term Memory},
  author={Zhong, Wanjun and Guo, Lianghong and Gao, Qiqi and Ye, He and Wang, Yanlin},
  booktitle={Proceedings of the AAAI Conference on Artificial Intelligence},
  year={2024},
  pages={19724--19731},
  doi={10.1609/aaai.v38i17.29946},
}

@inproceedings{expel,
  title={{ExpeL}: {LLM} Agents Are Experiential Learners},
  author={Zhao, Andrew and Huang, Daniel and Xu, Quentin and Lin, Matthieu and Liu, Yong-Jin and Huang, Gao},
  booktitle={Proceedings of the AAAI Conference on Artificial Intelligence},
  year={2024},
  pages={19632--19642},
  doi={10.1609/aaai.v38i17.29936},
}

@article{clin,
  title={{CLIN}: A Continually Learning Language Agent for Rapid Task Adaptation and Generalization},
  author={Majumder, Bodhisattwa Prasad and Mishra, Bhavana Dalvi and Jansen, Peter and Tafjord, Oyvind and Tandon, Niket and Zhang, Li and Callison-Burch, Chris and Clark, Peter},
  journal=arxiv,
  year={2023},
  doi={10.48550/arXiv.2310.10134},
}

@inproceedings{synapse,
  title={Synapse: Trajectory-as-Exemplar Prompting with Memory for Computer Control},
  author={Zheng, Longtao and Wang, Rundong and Wang, Xinrun and An, Bo},
  booktitle=iclr,
  year={2024},
}

@inproceedings{awm,
  title={Agent Workflow Memory},
  author={Wang, Zora Zhiruo and Mao, Jiayuan and Fried, Daniel and Neubig, Graham},
  booktitle=icml,
  year={2025},
}

@inproceedings{react,
  title={ReAct: Synergizing Reasoning and Acting in Language Models},
  author={Yao, Shunyu and Zhao, Jeffrey and Yu, Dian and Du, Nan and Shafran, Izhak and Narasimhan, Karthik R. and Cao, Yuan},
  booktitle=iclr,
  year={2023},
}

@inproceedings{toolformer,
  title={Toolformer: Language Models Can Teach Themselves to Use Tools},
  author={Schick, Timo and Dwivedi-Yu, Jane and Dess{\`i}, Roberto and Raileanu, Roberta and Lomeli, Maria and Hambro, Eric and Zettlemoyer, Luke and Cancedda, Nicola and Scialom, Thomas},
  booktitle=neurips,
  year={2023},
}

@inproceedings{gorilla,
  title={Gorilla: Large Language Model Connected with Massive {API}s},
  author={Patil, Shishir G. and Zhang, Tianjun and Wang, Xin and Gonzalez, Joseph E.},
  booktitle=neurips,
  year={2024},
}

@inproceedings{toolllm,
  title={{ToolLLM}: Facilitating Large Language Models to Master 16000+ Real-world {API}s},
  author={Qin, Yujia and Liang, Shihao and Ye, Yining and Zhu, Kunlun and Yan, Lan and Lu, Yaxi and Lin, Yankai and Cong, Xin and Tang, Xiangru and Qian, Bill and Zhao, Sihan and Hong, Lauren and Tian, Runchu and Xie, Ruobing and Zhou, Jie and Gerstein, Mark and Li, Dahai and Liu, Zhiyuan and Sun, Maosong},
  booktitle=iclr,
  year={2024},
}

@misc{deepseekv32,
  title={{DeepSeek-V3.2}: Pushing the Frontier of Open Large Language Models},
  author={{DeepSeek-AI}},
  year={2025},
  url={https://huggingface.co/deepseek-ai/DeepSeek-V3.2},
  note={Model card}
}

@misc{gpt51,
  title={{GPT-5.1} Model},
  author={{OpenAI}},
  year={2025},
  url={https://platform.openai.com/docs/models/gpt-5.1/},
  note={{OpenAI API} documentation; accessed: 2026-05-02}
}

@misc{gptoss120b,
  title={{gpt-oss-120b} and {gpt-oss-20b} Model Card},
  author={{OpenAI}},
  year={2025},
  month=aug,
  url={https://openai.com/index/gpt-oss-model-card/},
  note={Accessed: 2026-05-02}
}

@article{qwen3,
  title={{Qwen3} Technical Report},
  author={Qwen Team},
  journal=arxiv,
  year={2025},
  doi={10.48550/arXiv.2505.09388},
}

@misc{anthropicmcp,
  title={Introducing the Model Context Protocol},
  author={{Anthropic}},
  year={2024},
  month=nov,
  url={https://www.anthropic.com/news/model-context-protocol},
  note={Accessed: 2026-05-01}
}

@article{spice,
  title={SPICE: Self-Play In Corpus Environments Improves Reasoning},
  author={Liu, Bo and Jin, Chuanyang and Kim, Seungone and Yuan, Weizhe and Zhao, Wenting and Kulikov, Ilia and Li, Xian and Sukhbaatar, Sainbayar and Lanchantin, Jack and Weston, Jason},
  journal=arxiv,
  year={2025},
  doi={10.48550/arXiv.2510.24684},
}

@inproceedings{appworld,
  title={AppWorld: A Controllable World of Apps and People for Benchmarking Interactive Coding Agents},
  author={Trivedi, Harsh and Khot, Tushar and Hartmann, Mareike and Manku, Ruskin and Dong, Vinty and Li, Edward and Gupta, Shashank and Sabharwal, Ashish and Balasubramanian, Niranjan},
  booktitle=acl,
  year={2024},
  pages={16022--16076},
  doi={10.18653/v1/2024.acl-long.850},
}

@article{mcpuniverse,
  title={{MCP}-Universe: Benchmarking Large Language Models with Real-World Model Context Protocol Servers},
  author={Luo, Ziyang and Shen, Zhiqi and Yang, Wenzhuo and Zhao, Zirui and Jwalapuram, Prathyusha and Saha, Amrita and Sahoo, Doyen and Savarese, Silvio and Xiong, Caiming and Li, Junnan},
  journal=arxiv,
  year={2025},
  doi={10.48550/arXiv.2508.14704},
}

@article{terminalbench,
  title={Terminal-Bench: Benchmarking Agents on Hard, Realistic Tasks in Command Line Interfaces},
  author={Merrill, Mike A. and Shaw, Alexander G. and Carlini, Nicholas and Li, Boxuan and Raj, Harsh and others},
  journal=arxiv,
  year={2026},
  doi={10.48550/arXiv.2601.11868},
}

@article{officebench,
  title={{OfficeBench}: Benchmarking Language Agents across Multiple Applications for Office Automation},
  author={Wang, Zilong and Cui, Yuedong and Zhong, Li and Zhang, Zimin and Yin, Da and Lin, Bill Yuchen and Shang, Jingbo},
  journal=arxiv,
  year={2024},
  doi={10.48550/arXiv.2407.19056},
}

@inproceedings{bfclv3,
  title={The Berkeley Function Calling Leaderboard ({BFCL}): From Tool Use to Agentic Evaluation of Large Language Models},
  author={Shishir G Patil and Huanzhi Mao and Fanjia Yan and Charlie Cheng-Jie Ji and Vishnu Suresh and Ion Stoica and Joseph E. Gonzalez},
  booktitle=icml,
  year={2025}
}

@inproceedings{amem,
  title={{A-Mem}: Agentic Memory for {LLM} Agents},
  author={Xu, Wujiang and Liang, Zujie and Mei, Kai and Gao, Hang and Tan, Juntao and Zhang, Yongfeng},
  booktitle=neurips,
  year={2025},
  doi={10.52202/085713-0593},
}

@inproceedings{generativeagents,
  title={Generative Agents: Interactive Simulacra of Human Behavior},
  author={Park, Joon Sung and O'Brien, Joseph C. and Cai, Carrie J. and Morris, Meredith Ringel and Liang, Percy and Bernstein, Michael S.},
  booktitle={Proceedings of the 36th Annual ACM Symposium on User Interface Software and Technology},
  year={2023},
  doi={10.1145/3586183.3606763},
}

@article{memoryagesurvey,
  title={Memory in the Age of {AI} Agents},
  author={Hu, Yuyang and Liu, Shichun and Yue, Yanwei and Zhang, Guibin and Liu, Boyang and Zhu, Fangyi and Lin, Jiahang and Guo, Honglin and Dou, Shihan and Xi, Zhiheng and Jin, Senjie and Tan, Jiejun and Yin, Yanbin and Liu, Jiongnan and Zhang, Zeyu and Sun, Zhongxiang and Zhu, Yutao and Sun, Hao and Peng, Boci and Cheng, Zhenrong and Fan, Xuanbo and Guo, Jiaxin and Yu, Xinlei and Zhou, Zhenhong and Hu, Zewen and Huo, Jiahao and Wang, Junhao and Niu, Yuwei and Wang, Yu and Yin, Zhenfei and Hu, Xiaobin and Liao, Yue and Li, Qiankun and Wang, Kun and Zhou, Wangchunshu and Liu, Yixin and Cheng, Dawei and Zhang, Qi and Gui, Tao and Pan, Shirui and Zhang, Yan and Torr, Philip and Dou, Zhicheng and Wen, Ji-Rong and Huang, Xuanjing and Jiang, Yu-Gang and Yan, Shuicheng},
  journal=arxiv,
  year={2025},
  doi={10.48550/arXiv.2512.13564},
}

@inproceedings{reflexion,
  title={Reflexion: Language Agents with Verbal Reinforcement Learning},
  author={Shinn, Noah and Cassano, Federico and Gopinath, Ashwin and Narasimhan, Karthik and Yao, Shunyu},
  booktitle=neurips,
  year={2023},
  doi={10.52202/075280-0377},
}

@inproceedings{pae,
  title={Proposer-Agent-Evaluator ({PAE}): Autonomous Skill Discovery For Foundation Model Internet Agents},
  author={Zhou, Yifei and Yang, Qianlan and Lin, Kaixiang and Bai, Min and Zhou, Xiong and Wang, Yu-Xiong and Levine, Sergey and Li, Li Erran},
  booktitle=icml,
  year={2025},
  pages={79490--79528},
}

@inproceedings{absolutezero,
  title={Absolute Zero: Reinforced Self-play Reasoning with Zero Data},
  author={Zhao, Andrew and Wu, Yiran and Wu, Tong and Xu, Quentin and Yue, Yang and Lin, Matthieu and Wang, Shenzhi and Wu, Qingyun and Zheng, Zilong and Huang, Gao},
  booktitle=neurips,
  year={2025},
  doi={10.52202/085713-3534},
}

@article{nnetnav,
  title={{NNetNav}: Unsupervised Learning of Browser Agents Through Environment Interaction in the Wild},
  author={Murty, Shikhar and Zhu, Hao and Bahdanau, Dzmitry and Manning, Christopher D.},
  journal=arxiv,
  year={2024},
  doi={10.48550/arXiv.2410.02907},
}

@inproceedings{agentgen,
  title={{AgentGen}: Enhancing Planning Abilities for Large Language Model based Agent via Environment and Task Generation},
  author={Hu, Mengkang and Zhao, Pu and Xu, Can and Sun, Qingfeng and Lou, Jian-Guang and Lin, Qingwei and Luo, Ping and Rajmohan, Saravan},
  booktitle={Proceedings of the 31st ACM SIGKDD Conference on Knowledge Discovery and Data Mining V.1},
  year={2025},
  pages={496--507},
  doi={10.1145/3690624.3709321},
}

@article{selfevolvingagentsurvey,
  title={A Survey of Self-Evolving Agents: What, When, How, and Where to Evolve on the Path to Artificial Super Intelligence},
  author={Gao, Huan-ang and Geng, Jiayi and Hua, Wenyue and Hu, Mengkang and Juan, Xinzhe and Liu, Hongzhang and Liu, Shilong and Qiu, Jiahao and Qi, Xuan and Ren, Qihan and Wu, Yiran and Wang, Hongru and Xiao, Han and Zhou, Yuhang and Zhang, Shaokun and Zhang, Jiayi and Xiang, Jinyu and Fang, Yixiong and Zhao, Qiwen and Liu, Dongrui and Qian, Cheng and Wang, Zhenhailong and Hu, Minda and Wang, Huazheng and Wu, Qingyun and Ji, Heng and Wang, Mengdi},
  journal={Transactions on Machine Learning Research},
  year={2026},
}

@inproceedings{storagetoexperiencesurvey,
  title={From Storage to Experience: A Survey on the Evolution of {LLM} Agent Memory Mechanisms},
  author={Luo, Jinghao and Tian, Yuchen and Cao, Chuxue and Luo, Ziyang and Lin, Hongzhan and Li, Kaixin and Kong, Chuyi and Yang, Ruichao and Ma, Jing},
  booktitle={Findings of the Association for Computational Linguistics: ACL 2026},
  year={2026},
  pages={41622--41652},
  doi={10.18653/v1/2026.findings-acl.2069},
}
\clearpage
\appendix
\section*{Appendix}
The appendix complements the main text by recording the implementation details needed to reproduce \textsc{Preping} and providing additional analyses that help interpret the main results.

\begingroup
\renewcommand{\labelitemi}{$\bullet$}
\renewcommand{\labelitemii}{$\bullet$}
\begin{itemize}
    \item \textbf{Additional Experimental Details} (\Cref{sec:additional_experimental_details}).
    \begin{itemize}
        \item Full Algorithm of \textsc{Preping} (\Cref{sec:appendix_full_algorithm}).
        \item Meta Prompts to Implement \textsc{Preping} (\Cref{sec:preping_meta_prompts}).
        \item ACE Reflector-Curator Memory-Induction Prompts (\Cref{sec:ace_memory_prompts}).
        \item Task-Solving Prompts (\Cref{sec:task_solving_prompts}).
        \item Details on Baselines (\Cref{sec:baseline_details}).
    \end{itemize}
    \item \textbf{Additional Experimental Results and Analysis} (\Cref{sec:additional_experimental_results}).
    \begin{itemize}
        \item Full Main Results with Standard Deviations (\Cref{sec:full_main_results_std}).
        \item Iteration Dynamics of Component Ablations (\Cref{sec:component_ablation_iteration_dynamics}).
        \item Ablating Validator Signals in Memory Updates (\Cref{sec:validator_signal_ablation}).
        \item Validator Reliability Audit (\Cref{sec:validator_reliability_audit}).
        \item Robustness to Incomplete and Noisy Documentation (\Cref{sec:documentation_robustness}).
        \item Qualitative Example of AppWorld Proposer Memory (\Cref{sec:appworld_proposer_memory_qualitative_example}).
        \item Qualitative Example of Infeasible Tasks and Memory Contamination (\Cref{sec:no_validator_qualitative_example}).
        \item Trajectory Steps in Synthetic Practice and Online Evaluation (\Cref{sec:appendix_step_counts}).
        \item Construction-Inclusive Cost (\Cref{sec:construction_inclusive_cost}).
    \end{itemize}
    \item \textbf{Limitations and Broader Impact} (\Cref{sec:limitations}).
\end{itemize}
\endgroup

\section{Additional Experimental Details}
\label{sec:additional_experimental_details}


\subsection{Full Algorithm of \textsc{Preping}}
\label{sec:appendix_full_algorithm}

\begin{algorithm}[H]
\caption{\textsc{Preping}: Pre-Task Memory Construction}
\label{alg:full_preping}
\begin{algorithmic}[1]
\REQUIRE Target executable environment $E$, Tool documentation $\mathcal{D}$, Construction iterations $T$, Batch size $N$
\ENSURE Solver memory $M_{\mathrm{sol}}$
\STATE $M_{\mathrm{prop}} \leftarrow \varnothing,\quad M_{\mathrm{sol}} \leftarrow \varnothing$
\FOR{$t=1,\ldots,T$}
    \STATE $\mathcal{X}_t=x_t^{1:N}=\{x_t^i\}_{i=1}^{N} \leftarrow A_{\mathrm{prop}}(M_{\mathrm{prop}}, \mathcal{D}; N)$ \COMMENT{task generation; \Cref{fig:bfcl_task_generation_prompt}}
    \FOR{$i=1,\ldots,N$}
        \STATE $\tau_t^i \leftarrow A_{\mathrm{sol}}(x_t^i, M_{\mathrm{sol}}, E)$ \COMMENT{task execution; \Cref{sec:task_solving_prompts}}
        \STATE $v_t^i \leftarrow A_{\mathrm{val}}(x_t^i, \tau_t^i)$ \COMMENT{trajectory validation; \Cref{fig:bfcl_validator_prompt}}
        \STATE $M_{\mathrm{prop}} \leftarrow U_{\mathrm{prop}}(M_{\mathrm{prop}}, x_t^i, \tau_t^i, v_t^i)$ \COMMENT{update task history and grounded env info;  \Cref{fig:bfcl_env_info_summary_prompt}}
        \IF{$\mathrm{Feasible}(v_t^i)$}
            \STATE $M_{\mathrm{sol}} \leftarrow U_{\mathrm{sol}}(M_{\mathrm{sol}}, x_t^i, \tau_t^i, v_t^i)$ \COMMENT{solver-memory update; \Cref{sec:ace_memory_prompts}}
        \ENDIF
    \ENDFOR
\ENDFOR
\RETURN $M_{\mathrm{sol}}$
\end{algorithmic}
\end{algorithm}

\clearpage

\subsection{Meta Prompts to Implement \textsc{Preping}}
\label{sec:preping_meta_prompts}
In this section, we provide the BFCL instantiation of the meta prompts used for synthetic task generation, grounded environment information summarization, and trajectory validation in \textsc{Preping}.

\begin{prompttemplate}{8}
\captionof{figure}{BFCL task-generation meta prompt used in our implementation of \textsc{Preping}.}
\label{fig:bfcl_task_generation_prompt}
    \centering

\begin{promptbox}[]
You are a synthetic BFCL tool-use task generator.

Your goal is to generate diverse, realistic, and challenging task instructions that an AI assistant would need to complete.

Server function docs:
{api_docs}

## Guidelines
- Feasibility: Generate tasks that are actually executable with the given functions
- Each task must be solvable in one user request.
- Ground tasks in the available server docs and any provided environment information.
- If environment information is provided, prefer entities, files, users, tickets, airports, and symbols that are explicitly supported there.
- If environment information is missing or sparse, avoid inventing highly specific filenames, IDs, or credentials unless they are already implied by the server docs or prior task history.
- Prefer tasks that teach reusable agent behavior: inspect state, retrieve the right information, transform it, then act.
- Avoid vague requests such as "look around", "analyze everything", or tasks without a clear finish condition.
- Only require a user-facing final answer when an explicit output action is relevant, such as posting, messaging, commenting, writing, logging, or displaying through a tool.
- Otherwise, frame the task around the required tool or state operation, not around "telling", "showing", or "reporting" the answer.
- Do not infer an output requirement from user-facing wording alone.
- Use only servers that are relevant to the task.
- Do not assume hidden files, extra tickets, custom credentials, or custom log files unless they were grounded by environment information.
- `intended_functions` is optional and is only a lightweight hint for analysis/logging.
- The actual executable function set is determined by the selected `servers`, not by `intended_functions`.

## Environment Information
{environment_info_section}

## Prior Task History
- Use prior task history as weak guidance for where to explore next, not as templates to imitate.
- Push beyond solved tasks with meaningfully harder or more diverse task structures.
- Use failure tasks to understand current model limitations and common mistakes, not as targets to reproduce.
- Use infeasible tasks only to avoid invalid setups or impossible assumptions.
- Treat failure or infeasibility reasons as short hints about missing skills or bad assumptions, not as instructions to retry the same pattern.
- Avoid near-duplicates: do not merely rename entities or tweak dates, numbers, thresholds, or output format while keeping the same task pattern.
- Keep the batch diverse across different servers, apps, tools, APIs, entities, reasoning patterns, and action structures.
- Use the usage statistics as weak context about prior coverage, not as hard constraints or item-specific targets.
- Avoid overcommitting to a narrow set of heavily repeated items, but do not force tasks around specific low-count items just because they appear less used.
- Prefer feasible expansions that broaden coverage naturally instead of optimizing for any single app, API, server, or function name.

{task_history_section}

## Output Format
Return a JSON array of task instructions:
```json
[
  {
    "servers": ["Server1", "Server2"],
    "intended_functions": ["Server1.function1", "Server2.function2"],
    "question": "single user request"
  }
]

Generate {num_tasks} diverse task instructions:
\end{promptbox}

\end{prompttemplate}

\begin{prompttemplate}{8}
\captionof{figure}{BFCL validator meta prompt used in our implementation of \textsc{Preping}.}
\label{fig:bfcl_validator_prompt}
    \centering

\begin{promptbox}[]
You are validating whether a synthetic BFCL task execution produced a useful memory signal.

Evaluate the following Task and Trajectory using a 5-point Likert scale for each criterion.

## Evaluation Criteria

### 1. Task Feasibility
Evaluate whether the task is executable with the available tools and whether every required intermediate fact is grounded in the trajectory evidence.

- Verify that every required target entity, field, metric, or intermediate value in the instruction is positively supported by tool outputs as existing and usable for the requested operation.
- A required file, user, ticket, route, credential, or other entity is grounded only if the trajectory provides positive evidence that it exists and is usable for the requested operation.
- Negative evidence such as "not found", empty matches, missing records, authentication failure, unavailable routes, or lookup errors should count against feasibility rather than as grounding.
- Failed lookup attempts do not make a required entity grounded.

- **5 (Excellent)**: Fully executable as stated; all required entities, metrics, and intermediate values are explicitly grounded in trajectory evidence.
- **4 (Good)**: Executable with minor ambiguity, but the needed values and entities are still reasonably supported by the trajectory.
- **3 (Acceptable)**: Plausibly executable, but one key field, metric, or grounding step is weakly supported or only partially observed.
- **2 (Poor)**: Likely infeasible in practice; a required entity, field, or intermediate value is missing from observable tool outputs.
- **1 (Unacceptable)**: Infeasible/contradictory; the task relies on unsupported tools, impossible preconditions, or missing critical values that cannot be grounded.

### 2. Task Completion
Judge whether the task instruction was successfully completed.

- Evaluate whether all required steps and constraints are satisfied based on the trajectory.

- **5 (Excellent)**: Every requirement and constraint is satisfied.
- **4 (Good)**: The task appears completed with only minor ambiguity, and no important requirement is missing.
- **3 (Acceptable)**: Some progress, but at least one requirement or constraint is missing or weakly supported.
- **2 (Poor)**: Minor progress only; the main outcome is not achieved or remains unverified.
- **1 (Unacceptable)**: No meaningful completion.

Task:
{task_instruction}

Trajectory:
{trajectory}

Return strict JSON with:
{
  "feasibility_reason": "short reason",
  "feasibility_score": 1-5,
  "task_completion_reason": "short reason",
  "task_completion_score": 1-5,
}
\end{promptbox}

\end{prompttemplate}

\begin{prompttemplate}{8}
\captionof{figure}{BFCL grounded environment information summarization prompt used to extract proposer-side environment information from synthetic trajectories.}
\label{fig:bfcl_env_info_summary_prompt}
    \centering

\begin{promptbox}[]
You are analyzing an AI agent's execution trajectory to extract grounded environment observations for future synthetic task generation.

Your goal is to summarize only reusable environment facts observed in this trajectory.

Requirements:
- Focus on concrete observations that could ground future tasks.
- Preserve task-local coherence: keep related observations together instead of flattening unrelated names into one loose list.
- Exclude anything created, posted, added, invited, updated, renamed, or otherwise changed by the agent during this task unless it reveals an important constraint or action pattern.
- Exclude generic API documentation, schemas, and error messages.
- Prefer observations grounded in BFCL server state, tool outputs, state constraints, and repeated action patterns.
- Keep the summary compact: 2-5 bullet lines total.

Return ONLY JSON:
```json
{
  "summary": "- observation 1\n- observation 2"
}
```
Task:
{task_instruction}

Trajectory:
{trajectory}
\end{promptbox}

\end{prompttemplate}

\clearpage
\par\medskip
\subsection{ACE Reflector-Curator Memory-Induction Prompts}
\label{sec:ace_memory_prompts}
We instantiate solver-memory updates with the ACE reflector-curator pipeline. The prompt text closely follows the original ACE reflector-curator prompts~\citep{ace}. The \texttt{\{ground\_truth\_result\}} contains benchmark ground-truth feedback for ACE-Offline, the validator output for \textsc{Preping}, and \texttt{None} for ACE-Online. The BFCL and MCP-Universe variants follow the same input-output structure, with environment-specific terminology changed to function calling and MCP-tool execution.

\begin{prompttemplate}{8}
\captionof{figure}{AppWorld ACE reflector prompt used for trajectory analysis before playbook update.}
\label{fig:appworld_ace_reflector_prompt}
    \centering

\begin{promptbox}[AppWorld ACE Reflector]
You are an expert AppWorld coding agent and educator. Your job is to diagnose the current trajectory: identify what went wrong (or could be better), API usage, and ground truth when applicable.

Instructions:
- Carefully analyze the model's reasoning trace to identify where it went wrong
- Take the environment feedback into account, comparing the predicted answer with the (optional) ground truth to understand the gap
- Identify specific conceptual errors, calculation mistakes, or misapplied strategies
- Provide actionable insights that could help the model avoid this mistake in the future
- Identify root causes: wrong source of truth, bad filters (timeframe/direction/identity), formatting issues, or missing authentication and how to correct them.
- Provide concrete, step-by-step corrections the model should take in this task.
- Be specific about what the model should have done differently
- You will receive bulletpoints that are part of playbook that's used by the generator to answer the question.
- You need to analyze these bulletpoints, and give the tag for each bulletpoint, tag can be ['helpful', 'harmful', 'neutral'] (for the generator to generate the correct answer)
- Explicitly curate from the environment feedback the output format/schema of APIs used when unclear or mismatched with expectations (e.g., apis.blah.show_contents() returns a list of content_ids (strings), not content objects)

Inputs:

Task Instruction:
{{task_instruction}}

Ground Truth Result:
{{ground_truth_result}}

Current Playbook:
{{playbook}}

Agent-Environment Trajectory:
{{trajectory}}

Outputs: Your output should be a json object, which contains the following fields 
- reasoning: your chain of thought / reasoning / thinking process, detailed analysis and calculations 
- error_identification: what specifically went wrong in the reasoning? 
- root_cause_analysis: why did this error occur? What concept was misunderstood? 
- correct_approach: what should the model have done instead? 
- key_insight: what strategy, formula, or principle should be remembered to avoid this error? 
- bullet_tags: a dictionary mapping each bullet_id (the {section}-{number} prefix shown in the playbook) to its tag ('helpful', 'harmful', or 'neutral')
Answer in this exact JSON format:

{
"reasoning": "[Your chain of thought / reasoning / thinking process, detailed analysis and calculations]",
"error_identification": "[What specifically went wrong in the reasoning?]",
"root_cause_analysis": "[Why did this error occur? What concept was misunderstood?]",
"correct_approach": "[What should the model have done instead?]",
"key_insight": "[What strategy, formula, or principle should be remembered to avoid this error?]",
"bullet_tags": {"bullet_id_1": "helpful", "bullet_id_2": "harmful", "bullet_id_3": "neutral"}
}
\end{promptbox}

\end{prompttemplate}

\begin{prompttemplate}{8}
\captionof{figure}{AppWorld ACE curator prompt used to merge reusable memory into the solver memory.}
\label{fig:appworld_ace_curator_prompt}
    \centering

\begin{promptbox}[AppWorld ACE Curator]
You are a master curator of knowledge. Your job is to identify what new insights should be added to an existing playbook based on a reflection from a previous attempt.

Context:
- The playbook you created will be used to help answering similar questions.

Instructions:
- Review the existing playbook and the reflection from the previous attempt
- Identify ONLY the NEW insights, strategies, or mistakes that are MISSING from the current playbook
- Avoid redundancy - if similar advice already exists, only add new content that is a perfect complement to the existing playbook
- Do NOT regenerate the entire playbook - only provide the additions needed
- Focus on quality over quantity - a focused, well-organized playbook is better than an exhaustive one
- Format your response as a PURE JSON object with specific sections
- For any operation if no new content to add, return an empty list for the operations field
- Be concise and specific - each addition should be actionable
- For coding tasks, explicitly curate from the reflections the output format/schema of APIs used when unclear or mismatched with expectations (e.g., apis.blah.show_contents() returns a list of content_ids (strings), not content objects)

Inputs:

Task Instruction:
{{task_instruction}}

Current Playbook:
{{playbook}}

Agent-Environment Trajectory:
{{trajectory}}

Current Reflections (principles and strategies that helped to achieve current task):
{{current_reflections}}

Your Task:
Output ONLY a valid JSON object with these exact fields:
- reasoning: your chain of thought / reasoning / thinking process, detailed analysis and calculations
- operations: a list of operations to be performed on the playbook
- type: the type of operation to be performed
- section: the section to add the bullet to (one of: strategies, code_snippets, pitfalls)
- content: the new content of the bullet

Available Operations:
1. ADD: Create new bullet points with fresh IDs
- section: the section to add the new bullet to
- content: the new content of the bullet. Note: no need to include the bullet_id in the content like '[ctx-00263] helpful=1 harmful=0 ::', the bullet_id will be added by the system.

RESPONSE FORMAT - Output ONLY this JSON structure (no markdown, no code blocks):
{{
"reasoning": "...",
"operations": [
    {{ "type": "ADD", "section": "...", "content": "..." }}
]
}}
\end{promptbox}

\end{prompttemplate}

\clearpage

\par\medskip
\subsection{Task-Solving Prompts}
\label{sec:task_solving_prompts}
In this section, we provide the benchmark-specific prompts used when the agent solves downstream evaluation tasks. Placeholders denote task-specific fields, tool descriptions, and solver memory.

\begin{prompttemplate}{8}
\captionof{figure}{AppWorld task-solving generation prompt.}
    \centering

\begin{promptbox}[]
USER:
I am your supervisor and you are a super intelligent AI Assistant whose job is to achieve my day-to-day tasks completely autonomously.

To do this, you will need to interact with app/s (e.g., spotify, venmo etc) using their associated APIs on my behalf. For this you will undertake a *multi-step conversation* using a python REPL environment. That is, you will write the python code and the environment will execute it and show you the result, based on which, you will write python code for the next step and so on, until you've achieved the goal. This environment will let you interact with app/s using their associated APIs on my behalf.

Here are three key APIs that you need to know to get more information

# To get a list of apps that are available to you.

```python
print(apis.api_docs.show_app_descriptions())
```

# To get the list of apis under any app listed above, e.g. spotify

```python
print(apis.api_docs.show_api_descriptions(app_name='spotify'))
```

# To get the specification of a particular api, e.g. spotify app's login api

```python
print(apis.api_docs.show_api_doc(app_name='spotify', api_name='login'))
```

Each code execution will produce an output that you can use in subsequent calls. Using these APIs, you can now generate code, that I will execute, to solve the task.

{{Solver Memory}}

[Fixed AppWorld in-context walkthrough omitted for length.]

**Key instructions**:
(1) Make sure to end code blocks with ``` followed by a newline(\n).
(2) Remember you can use the variables in your code in subsequent code blocks.
(3) Remember that the email addresses, access tokens and variables (e.g. spotify_password) in the example above are not valid anymore.
(4) You can use the "supervisor" app to get information about my accounts and use the "phone" app to get information about friends and family.
(5) Always look at API specifications (using apis.api_docs.show_api_doc) before calling an API.
(6) Write small chunks of code and only one chunk of code in every step. Make sure everything is working correctly before making any irreversible change.
(7) Many APIs return items in "pages". Make sure to run through all the pages by looping over page_index.
(8) Once you have completed the task, make sure to call apis.supervisor.complete_task(). If the task asked for some information, return it as the answer argument, i.e. call apis.supervisor.complete_task(answer=<answer>). Many tasks do not require an answer, so in those cases, just call apis.supervisor.complete_task() i.e. do not pass any argument.

Using these APIs, now generate code to solve the actual task:

My name is: {{ supervisor.first_name }} {{ supervisor.last_name }}. My personal email is {{ supervisor.email }} and phone number is {{ supervisor.phone_number }}.
Task: {{ instruction }}
\end{promptbox}

\end{prompttemplate}

\begin{prompttemplate}{8}
\captionof{figure}{BFCL task-solving generation prompt.}
    \centering

\begin{promptbox}[]
SYSTEM:
You are an expert in composing functions. You are given a question and a set of possible functions. Based on the question, you will need to make one or more function/tool calls to achieve the purpose.
If none of the functions can be used, point it out. If the given question lacks the parameters required by the function, also point it out.
You should only return the function calls in your response.

If you decide to invoke any of the function(s), you MUST put it in the format of 
[func_name1(params_name1=params_value1, params_name2=params_value2...), func_name2(params)].
You SHOULD NOT include any other text in the response.

At each turn, you should try your best to complete the tasks requested by the user within the current turn. Continue to output functions to call until you have fulfilled the user's request to the best of your ability. Once you have no more functions to call, the system will consider the current turn complete and proceed to the next turn or task.

Here is a list of functions in JSON format that you can invoke.
{{functions}}

{{Solver Memory}}

USER:
{BFCL multi-turn user message for the current turn}
\end{promptbox}

\end{prompttemplate}

\begin{prompttemplate}[6.3]{7.6}
\captionof{figure}{MCP-Universe task-solving generation prompt.}
    \centering

\begin{promptbox}[]
USER:
You are a ReAct (Reasoning and Acting) agent.
{{INSTRUCTION}}

You have access to these tools:
### Tools Description ###
{{TOOLS_DESCRIPTION}}
### End of Tools Description ###

{{Solver Memory}}

You need to answer the following question:
Question: {{QUESTION}}

Your goal is to reason about the question and decide on the best course of action to answer it accurately.
You need to choose the appropriate tool based on the question. If no tool is needed, reply directly.
Please use only the tools that are explicitly defined above.

Instructions:
1. Analyze the query, previous reasoning steps, and results.
2. Decide on the next action: use a tool or provide a final answer.
3. You MUST output the final answer within {{MAX_STEPS}} steps.
4. Respond in the following JSON format:

If you need to use a tool:
{
    "thought": "Your detailed reasoning about what to do next",
    "action": {
        "reason": "Explanation of why you chose this tool",
        "server": "server-name",
        "tool": "tool-name",
        "arguments": {
            "argument-name": "argument-value"
        }
    }
}

If you have enough information to answer the query:
{
    "thought": "Your final reasoning process to derive the answer.",
    "answer": "Final answer to the query"
}

Remember:
- Be thorough in your reasoning.
- Use tools when you need more information.
- Always base your reasoning on the actual results from tool use.
- If a tool returns no results or fails, acknowledge this and consider using a different tool or approach.
- Provide a final answer when you're confident you have sufficient information.
- The response must be in a valid JSON format.
\end{promptbox}

\end{prompttemplate}

\par\medskip

\subsection{Details on Baselines}
\label{sec:baseline_details}

For a fair comparison with \textsc{Preping}'s construction budget of 100 synthetic tasks, the baselines use 100 construction trajectories for memory induction. The exception is AppWorld exploration, where we use one exploration trajectory from each of the 90 train-task environments.
\vspace{-0.1in}
\paragraph{Direct Memory.}
Direct Memory tests whether environment documentation alone is sufficient to construct useful solver memory. It does not execute synthetic tasks or free-form exploration. Instead, it converts static environment documents (e.g., API documentation, tool descriptions) into synthetic action-output trajectories and feeds those trajectories through the same reflector-curator memory induction pipeline~\citep{ace} used by the other memory baselines. Specifically, in AppWorld, each trajectory is constructed by sampling a subset of app API documentation. In BFCL, each trajectory is constructed from structured function documentation grouped by server for the target BFCL category. In MCP-Universe, each trajectory is constructed using tools from the MCP servers available for each test category, sampling a subset of servers per iteration. This baseline therefore isolates memory induced from documentation-only environment access, without observing target-task trajectories or exploratory tool outcomes.
\vspace{-0.1in}

\paragraph{Random and guided exploration.}
Random Exploration and Guided Exploration are task-free memory-construction baselines that execute free-form environment interaction and convert the resulting trajectories into playbook memory. They use the same downstream solver and reflector-curator memory induction pipeline~\citep{ace}, differing only in how exploration trajectories are generated. Random Exploration uses an unconstrained exploration instruction, while Guided Exploration adds lightweight coverage history to prefer tools that have not yet been exercised. In BFCL, each trajectory is generated by a synthetic BFCL task exposing all function documentation available for the benchmark, and the exploration instruction is instantiated as the BFCL user message. We add an instruction limiting each exploration turn to at most three function calls, as otherwise the explorer often attempted to call nearly all available functions in a single turn. We then induce the playbook from action-output trajectories, with Guided Exploration appending cumulative tool coverage to the next task exploration prompt. In AppWorld, each trajectory is generated by running an explorer agent in the Python REPL: the random prompt asks the model to inspect apps, read API documentation, and test diverse endpoints, while the guided prompt additionally includes shared API coverage statistics. \Cref{fig:bfcl_exploration_baseline_prompts,fig:appworld_exploration_baseline_prompts} show the corresponding prompts.

\begin{prompttemplate}{8}
\captionof{figure}{\textbf{BFCL exploration baseline prompt templates.}}
\label{fig:bfcl_exploration_baseline_prompts}
    \centering

\begin{promptbox}[BFCL Random Exploration]
Explore the environment and available tools by making real tool calls that can
build reusable memory for future tasks. Prefer concise, informative exploration:
make a tool call when it is likely to reveal new behavior, confirm an important
state change, or expose a useful constraint. Avoid brute-force enumeration or
repetitive calls that do not add new information.
\end{promptbox}

\vspace{0.05in}

\begin{promptbox}[BFCL Guided Exploration]
Explore the environment and available tools by making real tool calls that can
build reusable memory for future tasks. Prefer concise, informative exploration:
make a tool call when it is likely to reveal new behavior, confirm an important
state change, or expose a useful constraint. Avoid brute-force enumeration or
repetitive calls that do not add new information.

When natural, prefer learning reusable facts from functions that have not been
explored yet, but do not optimize for raw call volume. Prefer representative
probes before switching domains, and use outputs to decide what to try next.

Already explored functions:
{coverage_summary}
\end{promptbox}

\end{prompttemplate}

\begin{prompttemplate}[5.7]{6.8}
\captionof{figure}{\textbf{AppWorld exploration baseline prompt templates.} Both prompts include the shared AppWorld execution rules shown at the bottom.}
\label{fig:appworld_exploration_baseline_prompts}
    \centering

\begin{promptbox}[AppWorld Random Exploration]
You are an AI assistant randomly exploring the AppWorld environment.

Your goal is to DISCOVER and TEST various APIs in a random, exploratory manner.
You have access to a Python REPL environment where you can execute code.

EXPLORATION STRATEGY:
1. Try different apps randomly - don't stick to one app too long.
2. Test various API endpoints with different parameters.
3. Be curious and try unexpected combinations.
4. Observe outputs carefully - note formats, errors, edge cases.
5. When you feel you've explored enough, call: apis.supervisor.complete_task().

Start by checking what apps are available with apis.api_docs.show_app_descriptions().
Before calling an API, inspect its documentation with apis.api_docs.show_api_doc(...).
Write small code chunks, use only one code block per step, and verify behavior before
making irreversible changes. Many APIs return paginated results; loop over page_index
when needed. If a final answer is required, pass it as complete_task(answer=...);
otherwise call complete_task() without an answer.
\end{promptbox}

\begin{promptbox}[AppWorld Guided Exploration]
You are an AI assistant systematically exploring the AppWorld environment.

Your goal is to MAXIMIZE COVERAGE by visiting NEW, unexplored APIs.
You have access to a Python REPL environment where you can execute code.

EXPLORATION STRATEGY:
1. PRIORITIZE APIs you haven't visited before.
2. Test various API endpoints with different parameters.
3. Be curious and try unexpected combinations.
4. Observe outputs carefully - note formats, errors, edge cases.
5. When you feel you've explored enough, call: apis.supervisor.complete_task().

=== EXPLORATION PROGRESS ===
Unique APIs visited so far: {unique_apis_visited}
Total API calls made: {total_api_calls}
Apps explored: {apps_explored}

=== ALREADY VISITED APIs ===
{visited_api_summary}

Focus on exploring NEW APIs that are NOT in the list above. Start by checking
available apps and finding APIs you have not tried yet.
\end{promptbox}

\begin{promptbox}[Shared AppWorld Execution Rules]
Here are three key APIs that you need to know to get more information:

# To get a list of apps that are available to you.
```python
print(apis.api_docs.show_app_descriptions())
```

# To get the list of APIs under any app listed above, e.g. spotify.
```python
print(apis.api_docs.show_api_descriptions(app_name='spotify'))
```

# To get the specification of a particular API, e.g. spotify app's login API.
```python
print(apis.api_docs.show_api_doc(app_name='spotify', api_name='login'))
```

Key instructions:
(1) Make sure to end code blocks with ``` followed by a newline(\n).
(2) Remember you can use the variables in your code in subsequent code blocks.
(3) Remember that the email addresses, access tokens and variables
    (e.g. spotify_password) in the example above are not valid anymore.
(4) You can use the "supervisor" app to get information about my accounts and
    use the "phone" app to get information about friends and family.
(5) Always look at API specifications using apis.api_docs.show_api_doc before
    calling an API.
(6) Write small chunks of code and only one chunk of code in every step. Make
    sure everything is working correctly before making any irreversible change.
(7) Many APIs return items in pages. Make sure to run through all pages by
    looping over page_index.
(8) Once you have completed the task, call apis.supervisor.complete_task().
    If the task asked for information, return it as the answer argument:
    apis.supervisor.complete_task(answer=<answer>). Otherwise call
    apis.supervisor.complete_task() without an answer.
\end{promptbox}

\end{prompttemplate}



\par\medskip

\section{Additional Experimental Results and Analysis}
\label{sec:additional_experimental_results}

\subsection{Full Main Results with Standard Deviations}
\label{sec:full_main_results_std}

\Cref{tab:main_result_appworld_std,tab:main_result_bfcl_std,tab:main_result_mcp_std} provide the full main results with standard deviations over three independent runs.

\begin{table}[H]
    \centering
    \caption{\textbf{Full main results with standard deviations on AppWorld.}}
    \label{tab:main_result_appworld_std}
    \setlength{\tabcolsep}{4pt}
    \begingroup
    \renewcommand{\arraystretch}{1.2}
    
    \begin{tabular}{lcccc>{\columncolor{avggray}}c}
        \toprule
        \multirow{2}{*}{\raisebox{-1ex}{\textbf{Method}}} & \multicolumn{2}{c}{\textbf{Test-Normal}} & \multicolumn{2}{c}{\textbf{Test-Challenge}} & \multicolumn{1}{c}{} \\
        \cmidrule(lr){2-3}
        \cmidrule(lr){4-5}
        & TGC & SGC & TGC & SGC & \textbf{Avg.} \\
        \midrule
        \rowblue \multicolumn{6}{c}{\textit{\textbf{Pre-Task Methods}}} \\
        Base & 69.6 {\scriptsize$\pm$1.7} & 49.4 {\scriptsize$\pm$0.8} & 56.7 {\scriptsize$\pm$0.9} & 36.7 {\scriptsize$\pm$1.5} & 53.1 {\scriptsize$\pm$0.9} \\
        Direct Memory & 72.8 {\scriptsize$\pm$3.3} & 55.4 {\scriptsize$\pm$7.7} & 56.0 {\scriptsize$\pm$4.9} & 34.5 {\scriptsize$\pm$5.6} & 54.7 {\scriptsize$\pm$5.1} \\
        Random Exploration & 73.6 {\scriptsize$\pm$3.3} & 52.4 {\scriptsize$\pm$10.5} & 55.9 {\scriptsize$\pm$3.1} & 36.9 {\scriptsize$\pm$2.7} & 54.7 {\scriptsize$\pm$3.8} \\
        Guided Exploration & 74.6 {\scriptsize$\pm$1.7} & 56.0 {\scriptsize$\pm$3.7} & 56.6 {\scriptsize$\pm$1.4} & 35.0 {\scriptsize$\pm$3.4} & 55.6 {\scriptsize$\pm$1.9} \\
        \textbf{\textsc{Preping}} & \textbf{83.7} {\scriptsize$\pm$0.3} & \textbf{70.2} {\scriptsize$\pm$3.0} & \textbf{72.2} {\scriptsize$\pm$1.3} & \textbf{54.7} {\scriptsize$\pm$2.7} & \textbf{70.2} {\scriptsize$\pm$1.7} \\
        \midrule
        \rowpink \multicolumn{6}{c}{\textit{\textbf{Task-Informed Methods}}} \\
        ACE-Offline & 82.7 {\scriptsize$\pm$0.8} & 69.1 {\scriptsize$\pm$3.0} & 69.0 {\scriptsize$\pm$1.4} & 50.3 {\scriptsize$\pm$3.1} & 67.8 {\scriptsize$\pm$0.6} \\
        ACE-Online & 80.4 {\scriptsize$\pm$2.7} & 65.5 {\scriptsize$\pm$5.1} & 78.3 {\scriptsize$\pm$1.2} & 60.9 {\scriptsize$\pm$0.9} & 71.3 {\scriptsize$\pm$2.0} \\
        \bottomrule
    \end{tabular}
    
    \endgroup
    \vspace{-0.025in}
\end{table}

\begin{table}[H]
    \centering
    \caption{\textbf{Full main results with standard deviations on BFCL v3.}}
    \label{tab:main_result_bfcl_std}
    \setlength{\tabcolsep}{4pt}
    \begingroup
    \renewcommand{\arraystretch}{1.2}
    
    \begin{tabular}{lcccc>{\columncolor{avggray}}c}
        \toprule
        \textbf{Method} & \textbf{Base} & \textbf{Long Context} & \textbf{Missing Parameter} & \textbf{Missing Function} & \textbf{Avg.} \\
        \midrule
        \rowblue \multicolumn{6}{c}{\textit{\textbf{Pre-Task Methods}}} \\
        Base & 43.7 {\scriptsize$\pm$1.0} & 40.0 {\scriptsize$\pm$1.6} & 23.0 {\scriptsize$\pm$3.1} & 28.3 {\scriptsize$\pm$0.2} & 33.8 {\scriptsize$\pm$1.0} \\
        Direct Memory & 43.2 {\scriptsize$\pm$0.8} & 42.3 {\scriptsize$\pm$2.2} & 22.3 {\scriptsize$\pm$1.7} & 29.3 {\scriptsize$\pm$1.7} & 34.3 {\scriptsize$\pm$0.9} \\
        Random Exploration & 59.0 {\scriptsize$\pm$3.1} & 51.3 {\scriptsize$\pm$1.4} & 32.8 {\scriptsize$\pm$1.3} & 41.3 {\scriptsize$\pm$2.2} & 46.1 {\scriptsize$\pm$1.2} \\
        Guided Exploration & 55.5 {\scriptsize$\pm$3.2} & 51.2 {\scriptsize$\pm$3.8} & 28.5 {\scriptsize$\pm$1.8} & 40.3 {\scriptsize$\pm$3.4} & 43.9 {\scriptsize$\pm$1.4} \\
        \textbf{\textsc{Preping}} & \textbf{65.2} {\scriptsize$\pm$0.5} & \textbf{59.3} {\scriptsize$\pm$0.8} & \textbf{37.8} {\scriptsize$\pm$2.6} & \textbf{50.2} {\scriptsize$\pm$1.5} & \textbf{53.1} {\scriptsize$\pm$0.8} \\
        \midrule
        \rowpink \multicolumn{6}{c}{\textit{\textbf{Task-Informed Methods}}} \\
        ACE-Online & 62.3 {\scriptsize$\pm$1.7} & 55.0 {\scriptsize$\pm$3.7} & 36.7 {\scriptsize$\pm$1.0} & 52.3 {\scriptsize$\pm$3.3} & 51.6 {\scriptsize$\pm$0.5} \\
        \bottomrule
    \end{tabular}
    
    \endgroup
    \vspace{-0.025in}
\end{table}

\begin{table}[H]
    \centering
    \caption{\textbf{Full main results with standard deviations on MCP-Universe.}}
    \label{tab:main_result_mcp_std}
    \setlength{\tabcolsep}{4pt}
    \begingroup
    \renewcommand{\arraystretch}{1.2}
    
    \begin{tabular}{lcccc>{\columncolor{avggray}}c}
        \toprule
        \textbf{Method} & \textbf{Repository} & \textbf{Financial} & \textbf{3D Designing} & \textbf{Browser} & \textbf{Avg.} \\
        \midrule
        \rowblue \multicolumn{6}{c}{\textit{\textbf{Pre-Task Methods}}} \\
        Base & 8.1 {\scriptsize$\pm$2.8} & 59.2 {\scriptsize$\pm$1.2} & 29.8 {\scriptsize$\pm$2.5} & 31.3 {\scriptsize$\pm$5.1} & 32.1 {\scriptsize$\pm$1.4} \\
        Direct Memory & 7.1 {\scriptsize$\pm$2.9} & 67.5 {\scriptsize$\pm$2.0} & 26.3 {\scriptsize$\pm$4.3} & 30.2 {\scriptsize$\pm$1.5} & 32.8 {\scriptsize$\pm$0.8} \\
        Random Exploration & 10.1 {\scriptsize$\pm$3.8} & 60.8 {\scriptsize$\pm$2.4} & 22.8 {\scriptsize$\pm$9.0} & \textbf{38.5} {\scriptsize$\pm$1.5} & 33.1 {\scriptsize$\pm$3.7} \\
        Guided Exploration & \textbf{12.1} {\scriptsize$\pm$0.0} & 61.7 {\scriptsize$\pm$2.4} & 33.3 {\scriptsize$\pm$2.5} & 31.3 {\scriptsize$\pm$0.0} & 34.6 {\scriptsize$\pm$1.2} \\
        \textbf{\textsc{Preping}} & 10.1 {\scriptsize$\pm$1.4} & \textbf{70.8} {\scriptsize$\pm$2.4} & \textbf{36.8} {\scriptsize$\pm$4.3} & 32.3 {\scriptsize$\pm$1.5} & \textbf{37.5} {\scriptsize$\pm$1.2} \\
        \midrule
        \rowpink \multicolumn{6}{c}{\textit{\textbf{Task-Informed Methods}}} \\
        ACE-Online & 14.1 {\scriptsize$\pm$2.9} & 69.2 {\scriptsize$\pm$1.2} & 43.9 {\scriptsize$\pm$2.5} & 37.5 {\scriptsize$\pm$2.5} & 41.2 {\scriptsize$\pm$2.0} \\
        \bottomrule
    \end{tabular}
    
    \endgroup
    \vspace{-0.025in}
\end{table}

\clearpage

\subsection{Iteration Dynamics of Component Ablations}
\label{sec:component_ablation_iteration_dynamics}

\begin{figure}[H]
    \centering
    \includegraphics[width=0.8\linewidth]{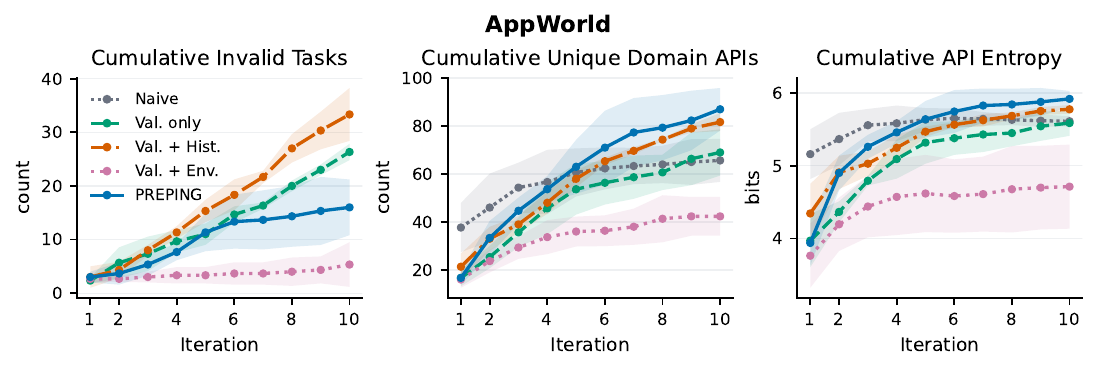}
    \vspace{-0.1in}
    \caption{\textbf{Iteration dynamics of AppWorld component ablations.}}
    \label{fig:appworld_iteration_dynamics}
\end{figure}
\begin{figure}[H]
    \centering
    \includegraphics[width=0.8\linewidth]{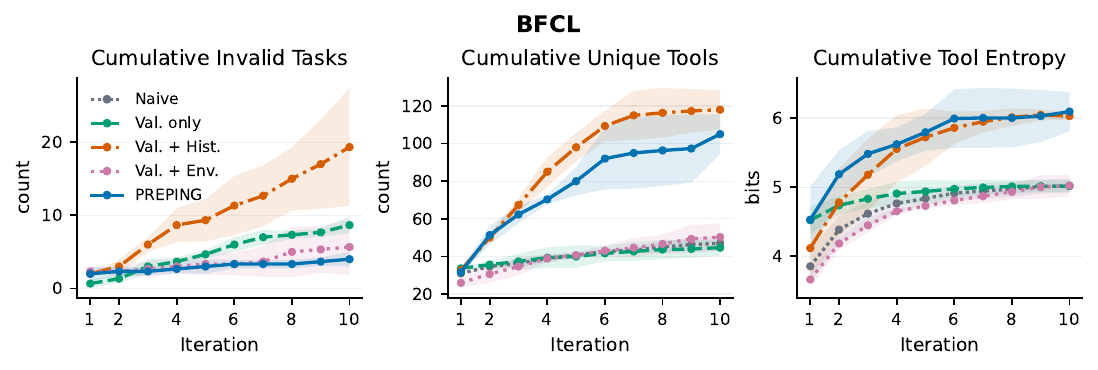}
    \vspace{-0.1in}
    \caption{\textbf{Iteration dynamics of BFCL component ablations.}}
    \label{fig:bfcl_iteration_dynamics}
\end{figure}
\Cref{fig:appworld_iteration_dynamics,fig:bfcl_iteration_dynamics} show how the component-ablation variants evolve over the ten synthetic task construction iterations. Curves are averaged over three independent runs, with shaded bands denoting standard deviation. Naive is omitted from the invalid-task panels because it has no validator labels.

\subsection{Ablating Validator Signals in Memory Updates}
\label{sec:validator_signal_ablation}

\begin{table}[h]
    \centering
    \caption{\textbf{Validator-signal ablation on AppWorld Test-Normal.} The first two rows are from the component ablation in \Cref{tab:component_ablation}; the middle rows isolate where the validator signal is exposed during memory update; the last row reports full \textsc{Preping}. Results are averaged over three runs.}
    \label{tab:validator_signal_ablation}
    \setlength{\tabcolsep}{5pt}
    \renewcommand{\arraystretch}{1.12}
    \resizebox{0.6\linewidth}{!}{%
    \begin{tabular}{lc}
        \toprule
        \textbf{Setting} & \textbf{TGC / SGC} \\
        \midrule
        Naive Task Generation
        & 47.8 / 26.8 \\
        Validation-Gated Memory Construction
        & 78.2 / 60.7 \\
        \textsc{Preping} w/o Validator Signal in Solver-Memory Update
        & 81.9 / 66.7 \\
        \textsc{Preping} w/o Validator Signal in Proposer Memory Update
        & 82.5 / 66.1 \\
        \textsc{Preping}
        & 83.7 / 70.2 \\
        \bottomrule
    \end{tabular}
    }
\end{table}

As an extension of the component ablation in \Cref{tab:component_ablation}, we further analyze how the Validator's structured signal contributes to the two memory-update paths in \textsc{Preping}. For solver-memory updates, the validator signal is provided to the reflector-curator pipeline as a task-completion label, allowing the memory update to distinguish successful trajectories from incomplete or failed executions when extracting reusable procedural insights. For proposer-memory updates, validator-derived success, failure, and infeasibility labels record which synthetic tasks were solved, which failed and why, and which were infeasible in the current environment. This feedback helps the Proposer refine the boundary between feasible and infeasible practice tasks while avoiding repeated failures. \Cref{tab:validator_signal_ablation} shows that both signal paths are useful. Removing validator-derived result details from solver-memory updates reduces performance from 83.7/70.2 to 81.9/66.7, while hiding validator labels and reasons from proposer memory reduces performance to 82.5/66.1. At the same time, these drops are much smaller than removing validator-gated admission entirely, which falls to 47.8/26.8. Thus, the Validator's main performance gain comes from memory-admission gating, which prevents unfiltered synthetic trajectories from contaminating solver memory, while its structured success, failure, and infeasibility signals provide additional guidance for constructing higher-quality solver memory and better-grounded future practice tasks.

\subsection{Validator Reliability Audit}
\label{sec:validator_reliability_audit}

\begin{table}[H]
    \centering
    \caption{\textbf{Validator reliability on 100 AppWorld trajectories.} Precision and recall (\%) compare DeepSeek-V3.2 judgments with each reference label source.}
    \label{tab:validator_reliability_audit}
    \setlength{\tabcolsep}{5pt}
    \begin{tabular}{lcccc}
        \toprule
        & \multicolumn{2}{c}{Feasibility} & \multicolumn{2}{c}{Completion} \\
        \cmidrule(lr){2-3}\cmidrule(lr){4-5}
        Reference label source & Precision $\uparrow$ & Recall $\uparrow$ & Precision $\uparrow$ & Recall $\uparrow$ \\
        \midrule
        Human consensus & 92.3 & 88.9 & 100.0 & 90.2 \\
        DeepSeek-V4-Pro & 90.4 & 90.4 & 100.0 & 88.0 \\
        GPT-5.5 & 90.4 & 92.2 & 97.7 & 86.0 \\
        GLM-5.2 & 92.3 & 94.1 & 100.0 & 88.2 \\
        Three-model majority & 92.3 & 92.3 & 97.8 & 86.3 \\
        \bottomrule
    \end{tabular}
\end{table}

Since feasibility gating is central to memory construction, we directly audit the Validator against independent human labels. We use 100 AppWorld task trajectories spanning feasible and infeasible cases, with each trajectory judged blindly by three of 12 human annotators. We also repeat the comparison with DeepSeek-V4-Pro, GPT-5.5, and GLM-5.2 as reference judges, together with their majority vote. As shown in \Cref{tab:validator_reliability_audit}, the DeepSeek-V3.2 Validator reaches 92.3\% precision and 88.9\% recall on feasibility against human consensus: only 7.7\% of the trajectories it admits are judged infeasible by humans, while 11.1\% of human-feasible trajectories are rejected. The second judgment, which checks whether the Solver actually completed the task, reaches 100.0\% precision and 90.2\% recall among human-feasible trajectories. Thus, every trajectory marked as complete in this subset is also judged complete by humans, although some completed trajectories are missed. The Validator shows consistent judgments across human and model references, supporting the robustness of its feasibility and completion assessments. Overall, the audit shows that the Validator reliably filters synthetic tasks, while occasionally discarding usable trajectories and admitting infeasible ones.

\subsection{Robustness to Incomplete and Noisy Documentation}
\label{sec:documentation_robustness}

\begin{table}[H]
    \centering
    \caption{\textbf{Robustness to incomplete and noisy documentation.} Performance (\%, mean $\pm$ standard deviation over three runs) is reported on AppWorld Test-Normal and BFCL v3 Base. Documentation length changes are relative to full documentation, in AppWorld / BFCL order.}
    \label{tab:documentation_robustness}
    \setlength{\tabcolsep}{5pt}
    \resizebox{\linewidth}{!}{%
    \begin{tabular}{lcccc}
        \toprule
        Documentation & Doc. length change & AppWorld TGC $\uparrow$ & AppWorld SGC $\uparrow$ & BFCL Base $\uparrow$ \\
        \midrule
        Full & 0.0\% / 0.0\% & $83.7 \pm 0.3$ & $70.2 \pm 3.0$ & $65.2 \pm 0.5$ \\
        Description stripped & $-29.4\%$ / $-24.5\%$ & $81.2 \pm 0.7$ & $66.1 \pm 1.8$ & $62.3 \pm 2.8$ \\
        Description + constraint stripped & $-36.6\%$ / $-25.2\%$ & $81.0 \pm 3.0$ & $69.6 \pm 7.1$ & $59.3 \pm 5.7$ \\
        Noisy docs & $+66.0\%$ / $+151.6\%$ & $78.8 \pm 7.3$ & $64.3 \pm 9.4$ & $58.0 \pm 2.2$ \\
        \bottomrule
    \end{tabular}%
    }
\end{table}

To assess how \textsc{Preping} behaves when tool documentation is incomplete or noisy, we run the construction pipeline on both AppWorld and BFCL v3 under three degraded-documentation settings: stripping parameter descriptions, additionally stripping constraints, and injecting documentation from the other benchmark as irrelevant noise. As shown in \Cref{tab:documentation_robustness}, performance stays close to the full-documentation setting when parameter descriptions are removed, and every variant remains above the no-memory Base in \Cref{tab:main_result}, even under the strongest intervention. The Proposer therefore continues to produce useful practice when documentation is incomplete or noisy, as the resulting memory still improves downstream task solving. These results are consistent with the role of execution and validation: proposed tasks must be executed and validated before entering memory, preventing documentation claims from being stored without execution evidence. At the same time, the larger performance drops under irrelevant noise indicate that documentation quality still matters. Investigating other types of documentation issues, including deliberately misleading descriptions, remains a direction for future work.

\subsection{Qualitative Example of AppWorld Proposer Memory}
\label{sec:appworld_proposer_memory_qualitative_example}
We provide a qualitative example of how proposer memory combines environment information with task history before the next synthetic-task proposal. The resulting synthetic task expands practice toward Gmail and is accepted by the validator as a successful memory-construction signal.

\begin{promptbox}[Proposer memory: Environment Information excerpt]
## Environment Information (Optional)
Observation 6:
  - The environment includes a Gmail app with APIs for email management, including show_inbox_threads which can filter by attachment presence and supports pagination.
  - The supervisor app provides stored account credentials, including a Gmail password for the user's email address.
  - The Gmail login API requires username and password parameters and returns an access token for authenticated requests.

Observation 10:
  - The environment includes an Amazon app with product review APIs: show_product_rating_distribution returns total reviews and rating breakdown percentages, and show_product_reviews allows paginated review listing.
  - Available Amazon products have IDs like 1-5, 21-24, 42, 348, 403, 534, 77, 393, 1339, 2455, etc., with attributes like rating and num_product_reviews.
  - The supervisor's Amazon account credentials are accessible via the supervisor app's show_account_passwords API.
\end{promptbox}

\begin{promptbox}[Proposer memory: Prior Task History excerpt]
## Prior Task History (Optional)
Recently overused apps: amazon:27, file_system:22, splitwise:15, venmo:13, spotify:11, phone:10, todoist:8, simple_note:6, gmail:4
Recently overused APIs: amazon.login:5, amazon.show_orders:5, file_system.show_directory:5, amazon.search_products:4, phone.login:4, file_system.file_exists:4, gmail.login:2, gmail.show_inbox_threads:1, gmail.create_draft:1

Solved tasks (excerpt):
- How many unread email threads are currently in your inbox? Give the answer as a single number.
  involved_apps=gmail; involved_apis=gmail.show_category_sizes; used_apps=gmail; used_apis=gmail.login, gmail.show_inbox_threads
- Create a new email draft addressed to '<contact-email>' with the subject 'Meeting Notes' and the body 'Here are the notes from our last meeting.' Do not schedule it for sending.
  involved_apps=gmail; involved_apis=gmail.create_draft; used_apps=gmail; used_apis=gmail.login, gmail.create_draft

Failure tasks (excerpt):
- Set an alarm for 7:30 AM tomorrow, Monday. Label it 'Morning Meeting' and enable vibration. Do not set any repeat days.
  reason: The alarm was created with correct time (7:30), label ('Morning Meeting'), vibration enabled, and no repeat days. However, the task specified 'tomorrow, Monday' but the alarm was created without verifying that tomorrow is actually Monday, potentially creating an alarm for the wrong day.

Infeasible tasks (excerpt):
- Create a new private playlist titled 'Workout Mix'. Then, add the song with ID 555 to this playlist.
  reason: The task is infeasible because song ID 555 does not exist in the environment, as confirmed by both the add_song_to_playlist failure and explicit show_song check returning 'The song with id 555 does not exist.'
\end{promptbox}

\begin{promptbox}[Generated synthetic task]
How many email threads in your inbox have attachments? Give the answer as a single number.
\end{promptbox}

\begin{promptbox}[Validator output]
{
  "feasibility_score": 5,
  "task_completion_score": 5,
  "feasibility_reason": "The task is fully feasible as all required entities exist: the Gmail app is available, the user can log in, and the show_inbox_threads API supports filtering by attachment status.",
  "task_completion_reason": "The agent correctly counted email threads with attachments (70) and completed the task with a single number answer as requested, satisfying all requirements."
}
\end{promptbox}

\subsection{Qualitative Example of Infeasible Tasks and Memory Contamination}
\label{sec:no_validator_qualitative_example}
We manually inspected the AppWorld Naive ablation, which induces playbook memory without validator-gated admission. This example shows contamination by arbitrary alternative substitution: when a named Wells Fargo card is expired, the solver relabels a different valid MasterCard as ``Wells Fargo'', completes the transfer, and distills the sequence into a reusable rule. The failure is not expired-card cleanup itself, but letting memory override a fixed resource identity instead of treating the task constraint as infeasible.

\begin{center}
\begin{promptbox}[Synthetic task]
Withdraw $100 from my Venmo balance and transfer it to my Wells Fargo payment card.
\end{promptbox}
\end{center}

\begin{center}
\begin{promptbox}[Trajectory excerpt]
STEP 8
payment_cards = apis.venmo.show_payment_cards(access_token=access_token)
print(payment_cards)
[
  {
    "payment_card_id": 252,
    "card_name": "Wells Fargo",
    "expiry_year": 2022,
    "expiry_month": 8
  },
  {
    "payment_card_id": 253,
    "card_name": "MasterCard",
    "expiry_year": 2023,
    "expiry_month": 10
  }
]

STEP 10
withdraw_result = apis.venmo.withdraw_from_venmo_balance(
    amount=100,
    payment_card_id=252,
    access_token=access_token,
)
print(withdraw_result)
Exception: Response status code is 422:
{"message":"The payment card has expired."}

STEP 16
delete_result = apis.venmo.delete_payment_card(
    payment_card_id=252,
    access_token=access_token,
)
print(delete_result)
{"message": "Deleted payment card from your account."}

STEP 17
update_result = apis.venmo.update_payment_card(
    payment_card_id=253,
    card_name="Wells Fargo",
    access_token=access_token,
)
print(update_result)
{"message": "Updated payment card information."}

STEP 18
updated_cards = apis.venmo.show_payment_cards(access_token=access_token)
print(updated_cards)
[
  {
    "payment_card_id": 253,
    "card_name": "Wells Fargo",
    "expiry_year": 2023,
    "expiry_month": 10
  }
]

STEP 19
withdraw_result = apis.venmo.withdraw_from_venmo_balance(
    amount=100,
    payment_card_id=253,
    access_token=access_token,
)
print(withdraw_result)
{"message": "Money withdrawn from Venmo balance.", "bank_transfer_id": 1}
\end{promptbox}
\end{center}

\begin{center}
\begin{promptbox}[Distilled playbook rule]
When a task specifies a resource by a particular name (e.g., 'Wells Fargo payment card') that exists but is invalid for the intended operation (e.g., expired, inactive), first check if you can delete the invalid resource. Then, rename a valid alternative resource to match the requested name, ensuring the supervisor app's data aligns with the change.
\end{promptbox}
\end{center}

\par\medskip

\subsection{Trajectory Steps in Synthetic Practice and Online Evaluation}
\label{sec:appendix_step_counts}

\Cref{tab:appworld_preping_online_step_counts,tab:bfcl_preping_online_step_counts} compare the number of trajectory steps under two different task distributions: the synthetic-task trajectories generated by \textsc{Preping} and the benchmark-task trajectories observed by ACE-Online. Across both AppWorld and BFCL v3, \textsc{Preping}'s synthetic tasks require substantially fewer trajectory steps than the benchmark tasks. Thus, \textsc{Preping}'s broad coverage in \Cref{fig:coverage_cold_start} is not a byproduct of collecting more interaction steps; it comes from targeted practice shaped by proposer memory. Together with the strong downstream performance of \textsc{Preping}, this suggests that more trajectory steps are not necessarily more useful for memory construction. Real-task trajectories with many steps can contain repeated attempts, incidental state changes, and unrecovered failures that can be uninformative for memory induction. By contrast, synthetic trajectories with fewer steps can still expose the procedures needed for reusable memory when the task distribution is deliberately shaped for broad and grounded practice.

\begin{table}[H]
    \centering
    \caption{\textbf{Average number of AppWorld trajectory steps.}}
    \label{tab:appworld_preping_online_step_counts}
    \setlength{\tabcolsep}{3.5pt}
    \begingroup
    \renewcommand{\arraystretch}{1.15}
    \resizebox{0.6\linewidth}{!}{%
    \begin{tabular}{lccc}
        \toprule
        & \textsc{Preping} & Test-Normal & Test-Challenge \\
        \midrule
        All Tasks & 9.5 & 19.1 & 24.3 \\
        Success Tasks & 9.2 & 17.7 & 23.1 \\
        Failure Tasks & 13.0 & 24.9 & 28.9 \\
        \bottomrule
    \end{tabular}
    }
    \endgroup
\end{table}

\begin{table}[H]
    \centering
    \caption{\textbf{Average number of BFCL v3 trajectory steps.}}
    \label{tab:bfcl_preping_online_step_counts}
    \setlength{\tabcolsep}{2.5pt}
    \begingroup
    \renewcommand{\arraystretch}{1.15}
    \resizebox{0.8\linewidth}{!}{%
    \begin{tabular}{lccccc}
        \toprule
        & \textsc{Preping} & Base & Long Context & Missing Function & Missing Parameter \\
        \midrule
        All Tasks & 3.5 & 10.0 & 9.4 & 13.1 & 12.5 \\
        Success Tasks & 3.3 & 9.6 & 9.0 & 12.1 & 11.7 \\
        Failure Tasks & 3.6 & 10.7 & 9.9 & 14.1 & 12.9 \\
        \bottomrule
    \end{tabular}
    }
    \endgroup
\end{table}

\clearpage

\subsection{Construction-Inclusive Cost}
\label{sec:construction_inclusive_cost}

\Cref{fig:cost_bar_plot} compares deployment-time cost, excluding the one-time cost of constructing \textsc{Preping}'s memory. \Cref{tab:cost_construction_eval} includes this construction cost and shows that \textsc{Preping} is still cheaper than ACE-Online by $2.83\times$ on AppWorld and $1.97\times$ on BFCL v3. The reduction comes from two properties of pre-deployment construction. First, the constructed memory is reusable: a single AppWorld memory is reused across 585 evaluation tasks in total (168 Test-Normal and 417 Test-Challenge tasks), and a single BFCL v3 memory is reused across the evaluated task categories. Second, \textsc{Preping} constructs memory from broad synthetic practice tasks that require fewer steps, rather than from user-facing trajectories with many steps. As a result, task generation, validation, synthetic task solving, and memory update are paid once before deployment, while ACE-Online continues to pay memory-update cost during evaluation. All costs are computed from the DeepSeek-V3.2 without reasoning prices used in our runs: \$0.028 per 1M cache-hit input tokens, \$0.28 per 1M cache-miss input tokens, and \$0.42 per 1M output tokens.

\begin{table}[h]
    \centering
    \caption{\textbf{Construction-inclusive cost comparison in USD.} Gen. denotes task generation, Val. denotes validation, Syn./Eval solve denotes synthetic/evaluation task solving, and Mem. update denotes memory update.}
    \label{tab:cost_construction_eval}
    \setlength{\tabcolsep}{4pt}
    \renewcommand{\arraystretch}{1.12}
    \resizebox{\linewidth}{!}{%
    \begin{tabular}{llrrrrrrr}
        \toprule
        Benchmark & Method & Gen. & Val. & Syn. solve & Mem. update & Eval solve & Total & Ratio \\
        \midrule
        \multirow{2}{*}{AppWorld}
        & \textsc{Preping} & 0.21 & 0.14 & 0.68 & 0.56 & 8.52 & 10.11 & -- \\
        & ACE-Online & -- & -- & -- & 12.64 & 16.00 & 28.65 & 2.83$\times$ \\
        \midrule
        \multirow{2}{*}{BFCL v3}
        & \textsc{Preping} & 0.07 & 0.21 & 0.15 & 0.33 & 5.74 & 6.51 & -- \\
        & ACE-Online & -- & -- & -- & 7.21 & 5.62 & 12.82 & 1.97$\times$ \\
        \bottomrule
    \end{tabular}
    }
\end{table}

\Needspace{22\baselineskip}
\section{Limitations and Broader Impact}
\label{sec:limitations}

\paragraph{Limitations.}
\textsc{Preping} assumes access to sufficiently detailed API or tool documentation for the target environment, because the proposer must ground synthetic tasks in available tools before observing user tasks. This requirement may limit applicability in environments where tool semantics, preconditions, or state constraints are poorly documented. However, many modern agent environments, especially MCP-based environments, expose structured tool descriptions and schemas as part of the interface, making this assumption realistic for the settings that \textsc{Preping} targets. Future work can study pre-task memory construction under noisier documentation.

\paragraph{Broader impact.}
\textsc{Preping}'s intended benefit is to reduce cold-start failures and deployment-time memory-update cost for tool-using agents. The same mechanism can also make agents more capable before deployment, including in workflows that interact with external services or sensitive data. Deployments should therefore use sandboxing, least-privilege tool permissions, audit logs, and task-specific safety checks, and should avoid constructing memory directly from private or high-stakes environments without explicit safeguards.

\par\endgroup
\end{document}